\documentclass{article}

\usepackage[final]{cpal_2025}

\usepackage{url}
\usepackage{amsmath}
\usepackage{amsfonts}
\usepackage{enumitem}
\usepackage{subcaption}
\usepackage{multirow}
\usepackage{graphicx}
\usepackage{algorithm}
\usepackage{algpseudocode}
\usepackage[table]{xcolor}
\usepackage{wrapfig}
\usepackage{booktabs}
\usepackage{euscript}

\title{Pruned Adaptation Modules: A Simple yet Strong Baseline for Continual Foundation Models}

\author{%
  Elif Ceren Gok Yildirim\textsuperscript{1}, ~Murat Onur Yildirim\textsuperscript{1}, ~Joaquin Vanschoren\textsuperscript{1} \\
  \textsuperscript{1}AMOR/e Lab, Eindhoven University of Technology \\
  \texttt{e.c.gok@tue.nl, m.o.yildirim@tue.nl, j.vanschoren@tue.nl}
}

\begin{document}

\maketitle

\begin{abstract}
The continual learning literature has rapidly shifted from traditional class-incremental learning (CIL) techniques to foundation model (FM)-based CIL methods without a clear understanding of how these newer approaches compare to strong, lightweight convolutional baselines. This abrupt transition has created a substantial methodological gap, making it difficult to assess whether recent FM-based CIL progress reflects genuine advances or merely the absence of rigorous baselines.
To address this gap, we introduce \emph{Pruned Adaptation Modules} (PAM), a simple yet effective method that freezes the vast majority of the pre-trained ResNet while enabling scalable continual adaptation through sparse task-specific layers. PAM yields up to a \textasciitilde5$\times$ reduction in trainable parameters and a \textasciitilde6$\times$ reduction in total parameters, significantly reducing the cost of continual updates. Across diverse benchmarks, PAM consistently mitigates catastrophic forgetting and outperforms state-of-the-art FM-based CIL approaches. Our findings position PAM as a strong and transparent baseline that helps bridge the gap between traditional and FM-based CIL, guiding future research for a more accurate assessment of true progress in continual adaptation.
\end{abstract}

\section{Introduction}

\begin{wrapfigure}[15]{r}{0.4\textwidth}
\captionsetup{font=small}
\centering
  \vspace{-22pt}
  \includegraphics[width=0.4\textwidth]{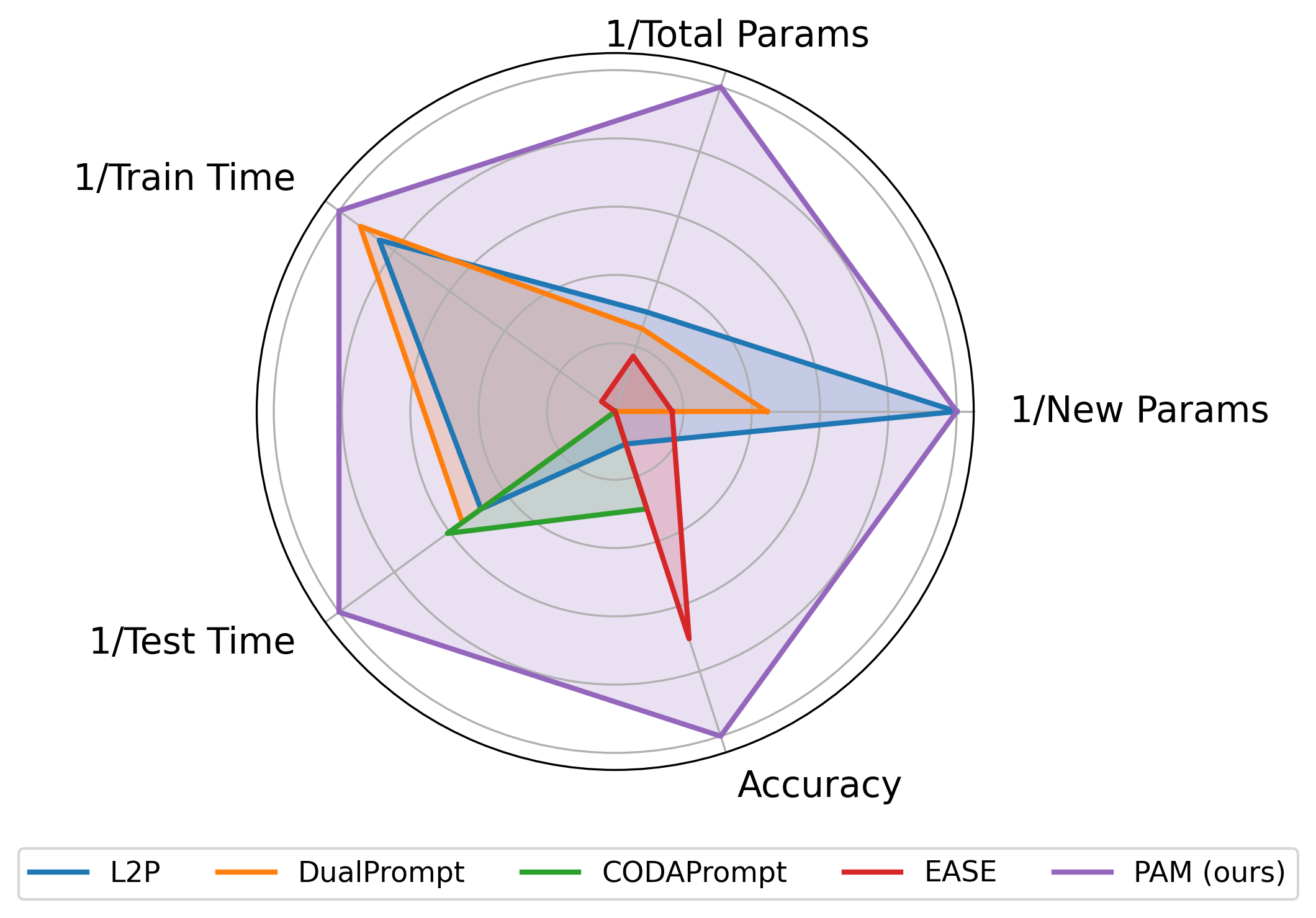}
  \vspace{-7pt}
  \caption{PAM is a simple yet powerful bridge that challenges the progress in FM–based CIL. It achieves better accuracy with ResNets, which significantly reduces runtime and parameters.}
  \label{fig:radar}
\end{wrapfigure}

Class-incremental learning (CIL) aims to enable models to acquire new knowledge over time without forgetting previously learned tasks. Recent work in this area has increasingly adopted foundation models (FMs), particularly Vision Transformers (ViTs)~\cite{vit}, due to their strong generalization properties and their demonstrated effectiveness as bases for incremental learning \cite{ptm_cil, pivot}. 

However, sequential fine-tuning of these models disrupts their pretrained representations, leading to pronounced catastrophic forgetting \cite{ranpac,zhang2023slca,panos2023first,zhou2024revisiting}.
To mitigate this, parameter-efficient fine-tuning (PEFT) strategies, such as prompt-based~\cite{codaprompt, sprompt, dualprompt, l2p} and adapter-based methods~\cite{simplecil, ease, mos}, freeze the FM and introduce small task-specific modules. These techniques restrict updates to a limited set of parameters, thereby enabling stabilized adaptation.

Despite their success, three key limitations remain. First, FM-based CIL methods typically rely on large-scale ViT backbones (e.g., 86M parameters), making training, storage, and deployment expensive in practical settings. Second, although designed to be lightweight, their task-specific modules are still relatively parameter-heavy; state-of-the-art prompt and adapter approaches require roughly \textasciitilde{3M} and \textasciitilde{1M} parameters per task, respectively. Third and most critically, the field has transitioned too quickly from traditional ConvNet-based CIL to FM-based CIL. As a result, it remains unclear whether these newer approaches truly outperform strong convolutional baselines. This rapid paradigm shift has created a methodological gap, obscuring the community’s ability to measure actual progress in continual learning with FMs.

To address this, we propose a novel PEFT-like approach that leverages pretrained ResNets \cite{resnet} to significantly reduce both the total and trainable parameters compared to existing FM-based techniques. Specifically, we freeze all layers except for the last one to leverage the transferable general features across tasks. Then, for each new task, we instantiate a dedicated last ResNet layer as a task-specific module, enabling efficient specialization in task-specific adaptation. To minimize both the total parameter count and the number of trainable parameters, we apply structured sparsity on these modules, referring to them as ‘Pruned Adaptation Modules’, or shortly PAM.

By incorporating these scalable design choices, we demonstrate a strong and principled bridging baseline that anchors future FM-based CIL research to a clear point of comparison, enabling the community to more reliably measure and advance true progress in continual learning.

\begin{figure}[t]
\captionsetup{font=small}
    \centering
    \includegraphics[width=0.85\textwidth]{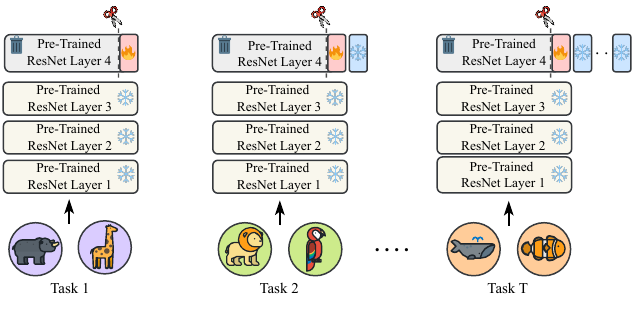} %0.75
    \caption{PAM freezes the first three layers of a pre-trained ResNet to preserve general knowledge while dynamically adding a task-specific last layer for each new task. To improve parameter efficiency, each last layer is structurally pruned to become ‘slim’ before training on its corresponding task. After training, the weights are frozen to prevent forgetting.}
    \label{fig:teaser}
  
\end{figure}

Our contributions are three-fold:

\begin{enumerate}[label=\Roman*., leftmargin=*, align=left]

\item We introduce ‘Pruned Adaptation Modules’ (PAM), a PEFT-inspired design for ConvNets, particularly ResNets, offering an alternative to existing FM-based approaches.

\item PAM achieves a $2$–$5\times$ reduction in trainable parameters and a $2$–$6\times$ reduction in total parameters compared to existing FM-based methods, enabling more efficient continual learning.

\item Across multiple benchmarks, PAM consistently outperforms adapter- and prompt-based methods, establishing itself as a simple yet strong baseline for evaluating and guiding future FM-based CIL research.

\end{enumerate}

%Specifically, our approach begins with a pre-trained ResNet model (e.g., ResNet18), where all layers are frozen, except the last feature extraction layer, to preserve the model’s pre-trained representations and maintain generalization across tasks. Then, for each task, We initialize a new last feature extraction layer as a task-specific module to adapt to given tasks efficiently. However, this task-specific block is modified to be highly parameter-efficient: we sparsify its weights aggressively, ensuring only a small subset of parameters remains trainable. By drastically reducing the active weights, our method minimizes both the total parameter and trainable parameter footprint while enabling task-specific adaptation efficiently. Importantly, instead of initializing the task-specific block randomly or using pre-trained weights, we leverage task similarity metrics to identify the most relevant prior task and initialize the new task-specific block with its corresponding weights. This informed initialization not only facilitates rapid convergence but also ensures the effective transfer of task-aligned knowledge. By focusing on ResNets, we demonstrate that competitive continual performance is achievable with significantly fewer parameters. This efficiency not only streamlines the learning process but also broadens the accessibility of continual learning for practical, real-world applications by offering lightweight architectures for efficient and sustainable continual learning.

\section{Related Work}
\paragraph{Traditional CIL.}
Traditional approaches seek to learn new classes sequentially from scratch while simultaneously mitigating catastrophic forgetting of previously acquired knowledge~\cite{threescenarios, yildirimceren, rebuffi2017icarl}. Existing approaches include rehearsal-based methods, which retain or synthesize exemplars from past tasks to balance data distributions during updates~\cite{rebuffi2017icarl, agem, liu2020mnemonics, zhao2021memory, aljundi2019gradient}; knowledge distillation-based techniques, leveraging teacher-student frameworks to align logits or features across incremental stages \cite{li2017learning, snell2017prototypical, hinton2015distilling, dhar2019learning, douillard2020podnet, zhang2020class}; and regularization-based strategies, which constrain parameter updates to preserve critical weights \cite{aljundi2018memory, aljundi2019task, rusu2016progressive, zenke2017continual}. Architectural adaptations address CIL through model rectification~\cite{pham2022continualnormalizationrethinkingbatch, shi2022mimicking, wu2019large, yu2020semantic, zhao2020maintaining} to correct decision biases or dynamic expansion~\cite{chen2023dynamic, douillard2022dytox, hu2023dense, huang2023resolving, yan2021dynamically, wang2022foster, cl-dst, serena} by incrementally adding task-specific networks.

\paragraph{FM-Based CIL.}
FM-based approaches have emerged as a powerful direction in CIL, enabling efficient adaptation to new classes by building upon the rich representations encoded in extensively pretrained models~\cite{mcdonnell2024premonition,wang2024hierarchical,zhou2024continual, tosca}. Recent works focus on PEFT methods which preserve the pretrained weights while integrating lightweight modules like prompts or adapters. 
L2P~\cite{l2p} adapts a technique from natural language processing by introducing a learnable prompt pool, where instance-specific prompts are selected via a key-query matching mechanism to guide the response of pre-trained models. 
DualPrompt~\cite{dualprompt} extends L2P by incorporating G-Prompt and E-Prompt, designed to capture task-invariant and task-specific information, respectively.
CODA-Prompt~\cite{codaprompt} employs contrastive loss to decorrelate prompt representations, mitigating interference, and integrates them using an attention-based weighting mechanism.
APER~\cite{zhou2024revisiting} systematically explores various PEFT methods, including adapters, and demonstrates that prototypical classifiers serve as a strong baseline. 
EASE~\cite{ease} enhances PTMs by attaching adapters to each layer which creates expandable subspaces and then aggregates feature representations from all adapter sets.

While these methods have advanced the state of the art, they are almost exclusively designed for and evaluated on large-scale pre-trained ViT backbones. However, the field's direct leap to these massive structures has left a significant gap in the literature regarding a simple but efficient baseline that we can truly assess their performance. 
\section{Method}
In this section, we first define the problem of CIL, outlining the challenge of sequentially learning from a stream of non-overlapping tasks. Next, we introduce our architectural design and module pruning. Finally, we detail our training and inference protocol, demonstrating how the model learns while dynamically selecting the appropriate module during evaluation without task identifiers. For a detailed description of PAM’s algorithm, please refer to the Appendix \ref{baseline-algo}.

\subsection{Problem Formulation}
CIL addresses the challenge of learning from a sequence of tasks 
$\mathcal{D}_1, \mathcal{D}_2, \dots, \mathcal{D}_B$, where each task 
$\mathcal{D}_b = \{(x_i, y_i)\}_{i=1}^{n_b}$ 
introduces a set of non-overlapping classes. Here, $x_i \in \mathbb{R}^D$ represents a training instance, $y_i\in Y_b$ denotes its corresponding label, and $Y_b \cap Y_{b'} = \emptyset$ for $b \neq b'$. During training on task $b$ the model has access only to $\mathcal{D}_b$, and the objective is two-fold: (i) acquiring new knowledge by learning to classify instances from the current task $\mathcal{D}_b$, and (ii) to preserve old knowledge by retaining performance on all previously seen tasks $\mathcal{D}_1, \dots, \mathcal{D}_{b-1}$.
After training on task $\mathcal{D}_b$, the model is evaluated on the cumulative label space $\mathcal{Y}_b = Y_1 \cup \dots \cup Y_b$. 
Specifically, the aim is to obtain a model $f(\mathbf{x}): X \rightarrow \mathcal{Y}_b$ that is able to classify all test dataset \textbf{without task indices} in the \textbf{exemplar-free setting}~\cite{l2p, dualprompt, codaprompt, sprompt, zhou2024revisiting, simplecil, ease}.

\subsection{Architecture Overview}
We build our method on the ResNets \cite{resnet}, and break the model down into three core components: a shared extractor $\Phi$, task-specific modules $\gamma$, and a unified classifier $W^\top$.
We keep the shared pre-trained extractor $\Phi: \mathbb{R}^D \rightarrow \mathbb{R}^d$ frozen throughout all learning sessions and serves as a generic task-invariant feature representation, allowing faster adaptation. Specifically, it corresponds to the initial three residual layers of a ResNet backbone.
For each task $\mathcal{D}_b$, a task-specific embedding module $\gamma_b: \mathbb{R}^d \rightarrow \mathbb{R}^d$ is appended to the shared frozen extractor $\Phi$. 
The unified classifier $W^\top: \mathbb{R}^d \rightarrow \mathbb{R}^{|\mathcal{Y}_b|}$ maps the final task-specific embeddings to class logits. Then, the overall model can be represented as given in Eq \ref{eq:logits}, where $\hat{y}_i$ produces the predictions with an activation function $\sigma$ for a given input $\mathbf{x_i}$ from task $b$.

%\vspace{-10pt}
\begin{equation}
\hat{y_i} = \arg\max \sigma(W^\top\gamma_b(\Phi(\mathbf{x}_i)))
\label{eq:logits}
\end{equation}

\subsection{Pruned Adaptation Modules}
We adopt a structured pruning strategy early in training to improve parameter efficiency. Specifically, after the first epoch, we evaluate the saliency of channels within each task-specific embedding module~$\gamma$ and remove the least informative ones. Consider a convolutional layer with weight tensor $W \in \mathbb{R}^{C_{\text{out}} \times C_{\text{in}} \times K \times K}$. For each output channel $c$, we compute its importance using the $L_1$-norm of its associated kernel weights. Formally, let $W_c \in \mathbb{R}^{C_{\text{in}} \times K \times K}$ denote the kernel corresponding to channel $c$, and let $W_c^i$ denote its individual weight values. The saliency score $s_c$ is defined as in Eq.~\ref{eq:l1norm}. Channels are then ranked according to their saliency $s_c$, and the lowest-scoring ones are pruned until the desired sparsity level is reached.
This structured pruning step substantially reduces the number of learnable parameters, ensuring that subsequent updates operate on a compact yet highly informative subset of weights. As a result, the method reduces parameter size while preserving sufficient expressive capacity for effective task adaptation.

\begin{equation}
s_c = \sum |W_c^i|.
\label{eq:l1norm}
\end{equation}

Following pruning, our architectural notation and flow are slightly modified. The shared pre-trained extractor $\Phi: \mathbb{R}^D \rightarrow \mathbb{R}^d$ remains unchanged, as does the unified classifier $W^\top: \mathbb{R}^d \rightarrow \mathbb{R}^{|\mathcal{Y}_b|}$. However, the task-specific module $\gamma_b$ is now replaced with a pruned adaptation module $\EuScript{S}_b$. The resulting architecture per task is then expressed as in Eq. \ref{eq:logits_sam}.

\begin{equation}
\hat{y_i} = \arg\max \sigma(W^\top\EuScript{S}_b(\Phi(\mathbf{x}_i))).
\label{eq:logits_sam}
\end{equation}

%\subsection{Task Similarity-Driven Initialization}
%We initialize each task-specific embedding module $\gamma_b$, with the weights of the most similar previously encountered task. To this end, we compute a task centroid $c_b$ by averaging the feature representations extracted from the shared frozen extractor $\Phi$ (i.e., the first three residual layers of a pretrained ResNet) over all training samples in \(\mathcal{D}_b\), assuming that the mean feature representation effectively captures the overall data distribution. We then measure the similarity between $c_b$ and the centroids of all previously encountered tasks, $\{c_1, \dots, c_{b-1}\}$, using the Manhattan distance as in Eq \ref{eq:centroid}. The task-specific block $\gamma_b$ for $\mathcal{D}_b$ is initialized by transferring the weights from the task that exhibits the smallest distance to $c_b$. This strategy leverages the most relevant prior knowledge to facilitate efficient transfer learning and robust adaptation to new task distributions.

%\begin{equation}
%d(\mathcal{D}_b, \mathcal{D}_i) \simeq \|c_b - c_i\|_1, \quad c_b = \frac{1}{n_b} \sum_{i=1}^{n_b} \Phi(x_i)
%\label{eq:centroid}
%\end{equation}

%\begin{equation}
%\gamma_b \leftarrow \gamma_{i^*}, \quad i^* = \underset{i \in \{1, \dots, b-1\}}{\arg\min} d(c_b, c_i)
%\label{eq:distance}
%\end{equation}

\subsection{Training and Inference Protocol}

For each learning session with dataset $\mathcal{D}_b$, we maintain the general feature extractor $\Phi$ in a frozen state to retain transferable and generalizable representations across tasks. This design choice ensures that the core knowledge learned from the pretraining remains intact while adapting to new tasks. Instead of modifying the entire network, we update only the parameters of the pruned task-specific module $\gamma$ and the shared classifier $W^\top$, enabling efficient adaptation with minimal interference between tasks. The parameters of these components are optimized using the standard cross-entropy loss, as defined in Eq. \ref{eq:celoss}.

\begin{equation}
\ell_{CE} = - \sum_{i=1}^{N} y_i \log(\hat{y}_i) =  - \sum_{i=1}^{N} y_i \log\sigma(W^\top\EuScript{S}_b(\Phi(\mathbf{x}_i)))
\label{eq:celoss}
\end{equation}

During inference, the model does not have access to task identities, necessitating a mechanism to select the most appropriate pruned task-specific module $\EuScript{S}$ for a given unlabeled test batch~$\mathbf{x}_{test}$. To address this challenge, we employ a confidence-based selection strategy. 
First, the test batch $\mathbf{x}_{test}$ is processed by the frozen general feature extractor $\Phi$, returning initial feature representations $\Phi(\mathbf{x}_{test})$. These representations are then passed through each pruned adaptation module $\EuScript{S}_b$ and the shared classifier $W^\top$, yielding task-conditioned class probability distributions as given in Eq \ref{eq:prob_dist}:  

\begin{equation}
p_b(x_{test}) = \sigma(W^\top\EuScript{S}_b(\Phi(\mathbf{x}_{test})))
\label{eq:prob_dist}
\end{equation}

Then, the confidence score is defined as the average maximum softmax probability across the batch, effectively capturing the certainty of each module in its predictions. Finally, the module with the highest confidence score is selected for inference, as formulated in Eq. \ref{eq:confidence}. This selection mechanism allows the model to dynamically adapt to different tasks without requiring explicit task identifiers, leveraging internal confidence measures to infer the most appropriate task-specific module.

\begin{equation}
\hat{b} = \arg\max_{b} \frac{1}{|\mathbf{x}_{test}|} \sum_{x_i \in \mathbf{x}_{test}} \max_{y \in \mathcal{Y}_b} p_b(y \mid x_i)
\label{eq:confidence}
\end{equation}

\section{Experimental Setup}

\paragraph{Datasets and Setting.}
We include two standard benchmarks CIFAR-100 and ImageNet-R as well as two fine-grained datasets CUB-200 and Cars-196. Specifically, CIFAR-100~\cite{cifar} comprises 100 natural image classes with 500 training images per class, while ImageNet-R~\cite{imagenetr} contains 200 classes with 24,000 training images and 6,000 test images. In addition, CUB-200~\cite{cub200} consists of 200 bird species, with approximately 60 images per class (equally divided between training and testing), and Cars-196~\cite{cars} includes 196 car models, with 8,144 training images and 8,040 test images. We adopt the ‘B-$m$ Inc-$n$’ notation to describe the class split, where $m$ denotes the number of classes in the initial stage and $n$ indicates the number of classes introduced at each incremental stage. These diverse datasets allow us to robustly evaluate our method across standard class-incremental learning scenarios and fine-grained classification tasks.

\paragraph{Comparison Methods.}
We benchmark our approach against state-of-the-art PTM-based class-incremental learning methods built on the ViT architecture. Specifically, we compare with SimpleCIL~\cite{simplecil}, L2P~\cite{l2p}, DualPrompt~\cite{dualprompt}, CODA-Prompt~\cite{codaprompt}, APER~\cite{simplecil}, and EASE~\cite{ease}. Additionally, we include sequential finetuning as a baseline to assess the impact of continual learning strategies.

\paragraph{Evaluation Metrics.} 
Following the benchmark protocol~\cite{ease}, we denote the model's accuracy after the $b-$th incremental stage as \(\mathcal{A}_b\). In our evaluation, we report both the final accuracy \(\mathcal{A}_B\) (i.e., the performance after the last stage) and the average accuracy across all stages, defined as \(\bar{\mathcal{A}} = \frac{1}{B} \sum_{b=1}^{B} A_b\), where \(B\) is the total number of incremental stages.

\paragraph{Implementation Details.} 
All experiments were conducted on an NVIDIA A100 GPU using PyTorch~\cite{pytorch} and the PILOT framework~\cite{pilot}. While existing methods utilize the pre-trained \textbf{ViT-B/16-IN1K} model which initially trained on ImageNet-21K and subsequently fine-tuned on ImageNet-1K, we employ pre-trained \textbf{ResNet18}, \textbf{ResNet50}, \textbf{ResNet101} and \textbf{ResNet152} models that are trained solely on ImageNet-1K. For our method, PAM, we train the models for 25 epochs using the Adam optimizer with a batch size of 48 and a learning rate of $0.001$. We perform the pruning immediately after the first epoch, enforcing a pruning magnitude of $96\%$, and continue training with the pruned module for the remaining epochs. To ensure fairness and robustness, we use the default parameters of existing approaches and repeat each experiment five times with different random seeds, where each seed also alters the task order. We report the mean and standard deviation of the performance metrics.  

\section{Results}
%In this section, we present a benchmark comparison of our method in terms of both performance and parameter efficiency. We then provide an ablation study followed by an in-depth analysis and discussion of our findings.

\subsection{State-of-art Comparison}

\paragraph{Incremental Performance.}
Table~\ref{tab:table1} presents the average and final accuracy of various continual learning methods across four benchmarks, demonstrating the effectiveness of PAM. We observe that on simpler datasets (e.g., CIFAR100), even a small ResNet18 backbone achieves competitive performance. For more challenging datasets such as CUB and ImageNet-R, larger ResNets yields improved results while still remaining smaller than ViT-B/16-IN21K used by other methods. These results establish PAM as a simple yet strong baseline for future FM-based continual learning, consistently outperforming existing approaches and providing a parameter-efficient, robust strategy for scalable continual adaptation.

\begin{figure}[t]
\captionsetup{font=small}
  \centering
  \begin{minipage}[t]{0.42\textwidth}
    \centering
    \includegraphics[width=\linewidth]{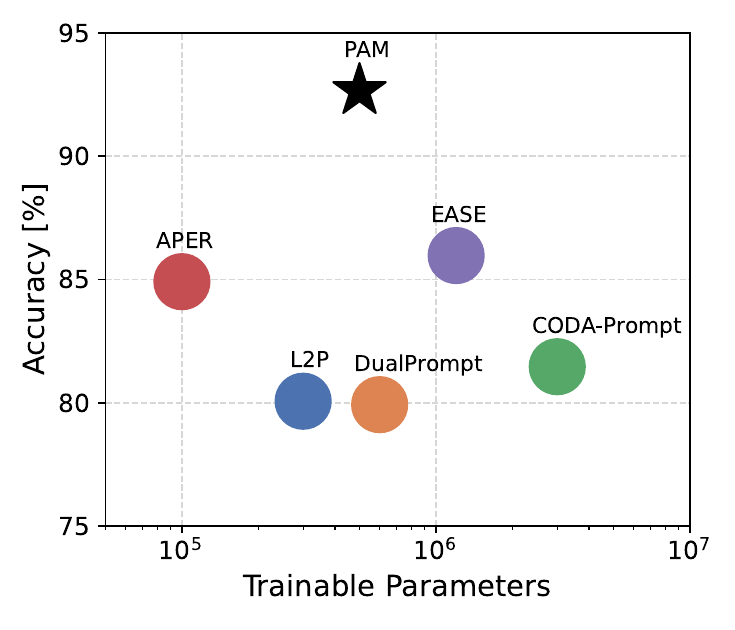}
    \label{fig:parameter_size}
  \end{minipage}\hspace{30pt}
  \begin{minipage}[t]{0.45\textwidth}
    \vspace{-147pt} % Adjust to move the table up
    \centering
    \renewcommand{\arraystretch}{2.8} % Increase row spacing
    \resizebox{\linewidth}{!}{%
      \begin{tabular}{lcccl}
        \toprule
        \LARGE Method & 
        \LARGE \shortstack{Trainable Params \\ Per Task} & 
        \LARGE \shortstack{Total Params \\ After All Tasks} & 
        \LARGE \shortstack{Final \\ Accuracy [\%]} \\
        \midrule
        \LARGE L2P         & \LARGE 300 K    & \LARGE 92 M   & \LARGE 80.06 \footnotesize{± 1.1} \\
        \LARGE DualPrompt  & \LARGE 600 K   & \LARGE 98 M   & \LARGE 79.92 \footnotesize{± 0.4} \\
        \LARGE CODA-Prompt & \LARGE 3 M   & \LARGE 146 M  & \LARGE 81.46 \footnotesize{± 0.3} \\
        \LARGE APER        & \LARGE 100 K   & \LARGE 86 M   & \LARGE 84.91 \footnotesize{± 0.2} \\
        \LARGE EASE        & \LARGE 1.2 M   & \LARGE 110 M  & \LARGE 85.97 \footnotesize{± 0.6} \\ \hline
        \rowcolor{pink!20}
        \LARGE PAM (RN18)       & \LARGE 600 K  & \LARGE 15 M  & \LARGE 88.51 \footnotesize{± 3.4} \\
        \rowcolor{pink!20}
        \LARGE PAM (RN50)      & \LARGE 600 K  & \LARGE 21 M  & \LARGE 92.50 \footnotesize{± 2.1} \\
        \rowcolor{pink!20}
        \LARGE PAM (RN101)      & \LARGE 600 K  & \LARGE 40 M  & \LARGE 93.05 \footnotesize{± 1.7} \\
        \rowcolor{pink!20}
        \LARGE PAM (RN152)      & \LARGE 600 K  & \LARGE 56 M  & \LARGE 93.79 \footnotesize{± 1.7} \\
        \bottomrule
      \end{tabular}
    }
    \label{tab:parameter_tab}

  \end{minipage}
  \vskip -0.7cm
  \caption{Parameter size vs. accuracy: The left panel shows that PAM challenges existing and future FM-based methods; and the right panel presents the parameter count for different methods after completing all sessions.}
  \label{fig:parameter_fig}
\end{figure}

\paragraph{Parameter Size Comparison.}
Figure~\ref{fig:parameter_fig} compares FM-based CIL methods on the CIFAR B0 Inc5 benchmark in terms of accuracy and parameter efficiency. PAM uses $5\times$ and $2\times$ fewer trainable parameters than the state-of-the-art prompt-based CODA-Prompt and adapter-based EASE, respectively, while delivering superior performance. Considering total parameters, including the frozen pre-trained backbone and task-specific modules, PAM requires $2$–$6\times$ fewer parameters than these baselines. These results show that ResNet models can perform just as well, suggesting that current FM-based CIL methods are not fully leveraging their potential.

\begin{table}[t]
\captionsetup{font=small}
\caption{Average and final accuracy [\%] with methods using \textbf{ViT-B/16-IN21K} and PAM using \textbf{ResNets}.}
\label{tab:table1}
\fontsize{11}{14}\selectfont
\resizebox{\textwidth}{!}{%
\begin{tabular}{lcccccccc}
\hline
\multirow{2}{*}{Method} &
  \multicolumn{2}{c}{CIFAR B0 Inc5} &
  \multicolumn{2}{c}{CUB B0 Inc10} &
  \multicolumn{2}{c}{IN-R B0 Inc20} &
  \multicolumn{2}{c}{Cars B0 Inc10} \\
 &
  $\bar{\mathcal{A}}$ &
  $\mathcal{A}_B$ &
  $\bar{\mathcal{A}}$ &
  $\mathcal{A}_B$ &
  $\bar{\mathcal{A}}$ &
  $\mathcal{A}_B$ &
  $\bar{\mathcal{A}}$ &
  $\mathcal{A}_B$ \\ \hline
Finetune &
  60.65 \footnotesize{± 5.6} &
  48.12 \footnotesize{± 3.3} &
  55.78 \footnotesize{± 2.8} &
  33.13 \footnotesize{± 3.3} &
  59.09 \footnotesize{± 3.7} &
  49.46 \footnotesize{± 3.3} &
  41.90 \footnotesize{± 1.0} &
  19.47 \footnotesize{± 2.7} \\
SimpleCIL &
  86.48 \footnotesize{± 0.8} &
  81.28 \footnotesize{± 0.1} &
  91.58 \footnotesize{± 1.3} &
  86.73 \footnotesize{± 0.1} &
  61.31 \footnotesize{± 0.4} &
  54.55 \footnotesize{± 0.1} &
  54.95 \footnotesize{± 0.8} &
  35.43 \footnotesize{± 0.0} \\
L2P &
  84.90 \footnotesize{± 1.2} &
  80.06 \footnotesize{± 1.4} &
  73.22 \footnotesize{± 1.8} &
  61.55 \footnotesize{± 1.7} &
  75.92 \footnotesize{± 0.7} &
  70.88 \footnotesize{± 0.7} &
  42.06 \footnotesize{± 2.0} &
  30.07 \footnotesize{± 0.8} \\
DualPrompt &
  85.61 \footnotesize{± 1.3} &
  79.92 \footnotesize{± 0.4} &
  81.36 \footnotesize{± 1.8} &
  70.51 \footnotesize{± 1.1} &
  71.48 \footnotesize{± 0.5} &
  66.09 \footnotesize{± 1.3} &
  45.30 \footnotesize{± 1.1} &
  30.15 \footnotesize{± 0.9} \\
CODA-Prompt &
  87.64 \footnotesize{± 0.4} &
  81.46 \footnotesize{± 0.3} &
  77.65 \footnotesize{± 1.0} &
  68.44 \footnotesize{± 1.0} &
  76.25 \footnotesize{± 0.3} &
  71.39 \footnotesize{± 0.3} &
  36.22 \footnotesize{± 0.6} &
  25.44 \footnotesize{± 0.3} \\
APER-Adapter &
  89.57 \footnotesize{± 0.9} &
  84.91 \footnotesize{± 0.2} &
  91.62 \footnotesize{± 1.2} &
  86.72 \footnotesize{± 0.2} &
  74.81 \footnotesize{± 0.8} &
  66.97 \footnotesize{± 0.8} &
  47.91 \footnotesize{± 0.8} &
  35.49 \footnotesize{± 0.0} \\
EASE &
  90.79 \footnotesize{± 0.8} &
  85.97 \footnotesize{± 0.6} &
  \textbf{92.51} \footnotesize{± 1.3} &
  86.49 \footnotesize{± 1.2} &
  \textbf{80.35} \footnotesize{± 1.0} &
  75.74 \footnotesize{± 0.8} &
  49.32 \footnotesize{± 1.0} &
  34.75 \footnotesize{± 0.3} \\ \hline
\rowcolor{pink!20}
 PAM (RN18)  &
  \textbf{91.40} \footnotesize{± 2.1} &
  \textbf{88.51} \footnotesize{± 3.4} &
  87.40 \footnotesize{± 1.3} &
  83.69 \footnotesize{± 3.4} &
  68.54 \footnotesize{± 0.4} &
  65.76 \footnotesize{± 0.4} &
  \textbf{79.09} \footnotesize{± 1.6} &
  \textbf{64.82} \footnotesize{± 1.6} \\ 
\rowcolor{pink!20}
PAM (RN50)  &
  \textbf{93.06} \footnotesize{± 1.5} &
  \textbf{92.50} \footnotesize{± 2.1} &
  85.40 \footnotesize{± 3.0} &
  82.67 \footnotesize{± 3.0} &
  73.22 \footnotesize{± 0.6} &
  72.83 \footnotesize{± 0.5} &
  \textbf{77.41} \footnotesize{± 1.5} &
  \textbf{62.23} \footnotesize{± 8.2} \\ 
\rowcolor{pink!20}
 PAM (RN101)  &
  \textbf{94.16} \footnotesize{± 1.5} &
  \textbf{93.05} \footnotesize{± 1.7} &
  89.76 \footnotesize{± 1.1} &
  \textbf{87.26} \footnotesize{± 1.7} &
  77.75 \footnotesize{± 0.7} &
  \textbf{77.03} \footnotesize{± 0.8} &
  \textbf{80.16} \footnotesize{± 2.1} &
  \textbf{77.30} \footnotesize{± 2.6} \\ 
\rowcolor{pink!20}
PAM (RN152)  &
  \textbf{94.17} \footnotesize{± 1.4} &
  \textbf{93.79} \footnotesize{± 1.7} &
  89.91 \footnotesize{± 1.4} &
  \textbf{88.35} \footnotesize{± 1.5} &
  79.33 \footnotesize{± 1.0} &
   \textbf{78.95} \footnotesize{± 0.5} &
  \textbf{83.10} \footnotesize{± 0.9} &
  \textbf{77.23} \footnotesize{± 3.5} \\ \hline
\end{tabular}%
}
%\vskip -0.1cm
\end{table}

\subsection{Ablation Study}
To systematically evaluate the impact of core design choices for PAM, we conduct ablations across three critical dimensions: pruning schedule, pruning magnitude, and pruned adaptation module selection strategies during inference. Furthermore, we also evaluate knowledge transfer between modules to see if `warm-starting' is beneficial.

\begin{figure}[t]
\captionsetup{font=small}
    \centering
    \begin{subfigure}{0.325\textwidth}
        \centering
        \includegraphics[width=\textwidth]{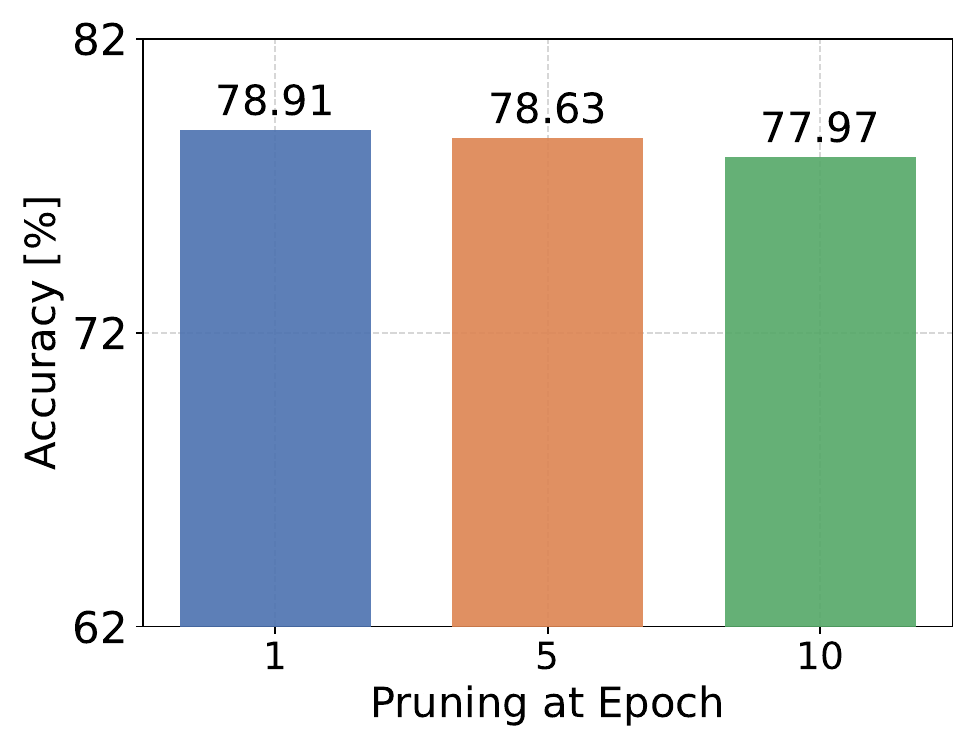}
        \caption{Pruning schedule}
        \label{ablation_pruning_epoch}
    \end{subfigure}
    \hfill
    \begin{subfigure}{0.325\textwidth}
        \centering
        \includegraphics[width=\textwidth]{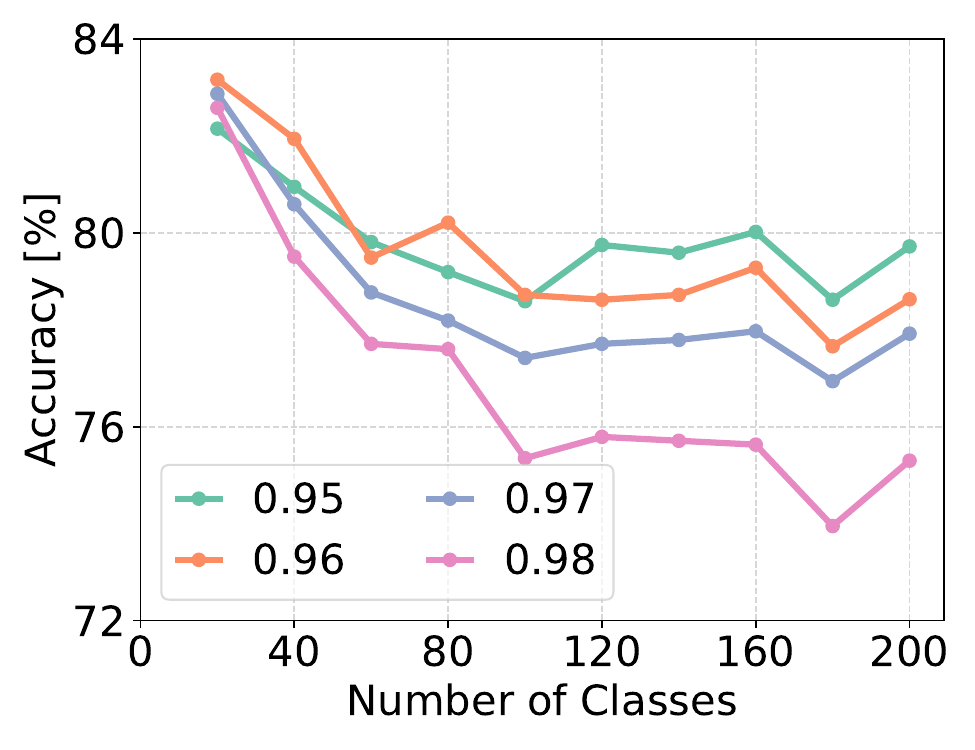}
        \caption{Pruning magnitude}
        \label{ablation_pruning}
    \end{subfigure}
    \hfill
    \begin{subfigure}{0.325\textwidth}
        \centering
        \includegraphics[width=\textwidth]{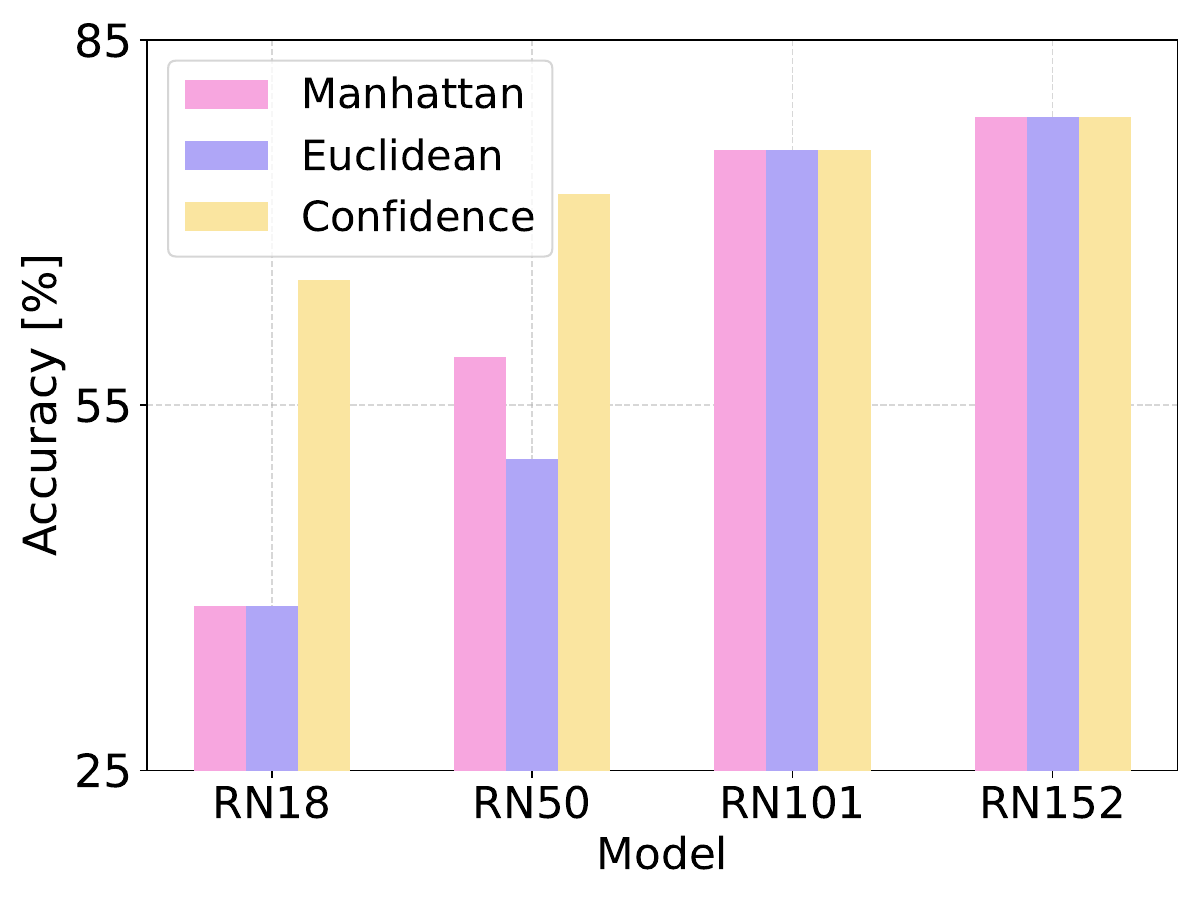}
        \caption{Inference time module selection}
        \label{ablation_test_time}
    \end{subfigure}
    \caption{Ablations of different components for the PAM method. (a) Effect of pruning timing on the performance. (b) Impact of different sparsity levels on performance. (c) Comparison of task-specific adaptation module selection strategies during inference.}
    \vskip -0.3cm
  \end{figure}

\paragraph{Pruning schedule.} We investigate the impact of \textit{‘when to prune’} and evaluate three scenarios where pruning is performed after the $1^{\text{st}}$, $5^{\text{th}}$, and $10^{\text{th}}$ training epoch, respectively. As shown in Figure~\ref{ablation_pruning_epoch}, these experiments reveal the sensitivity of the model's performance to the timing of pruning, thereby informing the optimal schedule. Our observations indicate that applying pruning to the task-specific module in the early stages of training is more beneficial than applying it in later epochs,  thereby motivating us to implement module pruning at epoch $1$.

\begin{figure}[t]
\captionsetup{font=small}
    \centering
    \begin{subfigure}{0.315\textwidth}
        \centering
        \includegraphics[width=\textwidth]{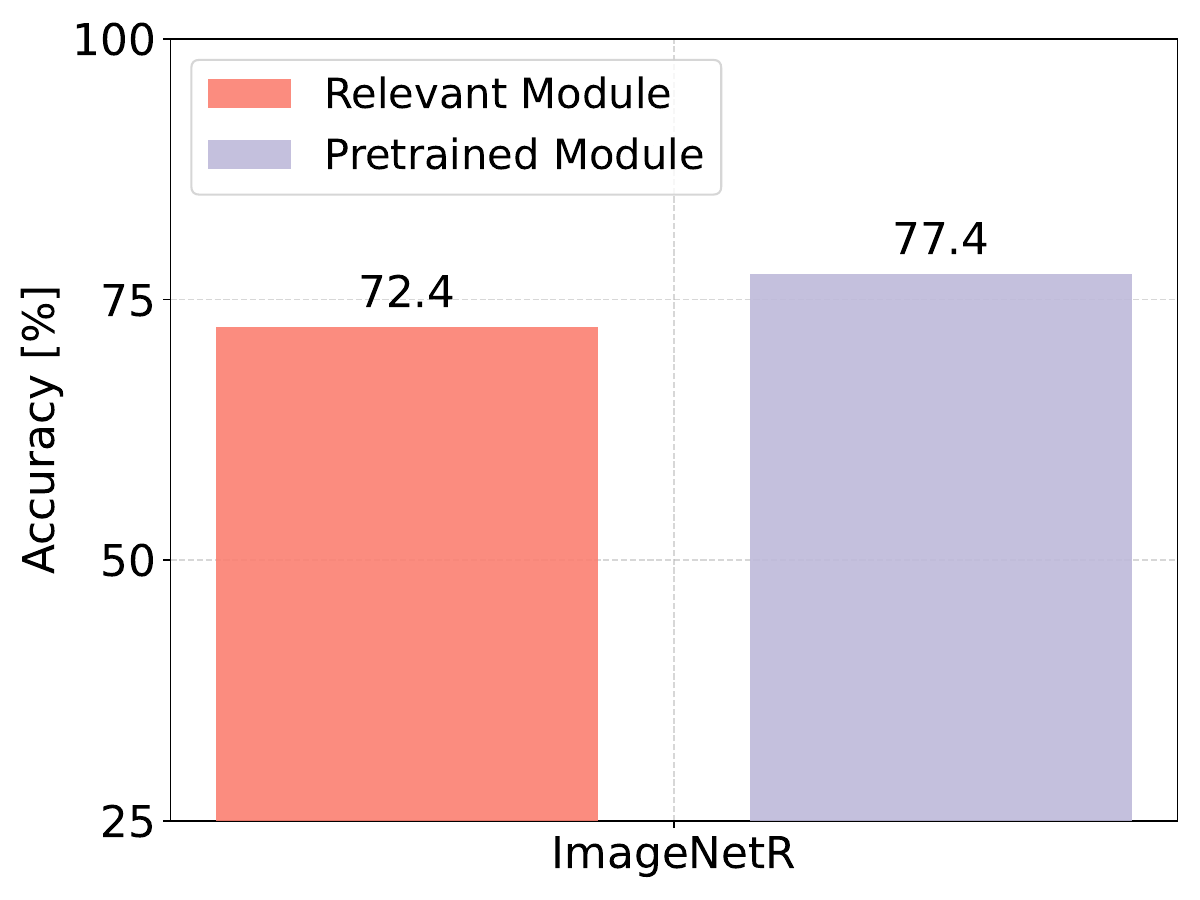}
        \caption{Final incremental accuracy}
        \label{knowledge_transfer_barplot}
    \end{subfigure}
    \hfill
    \begin{subfigure}{0.32\textwidth}
        \centering
        \includegraphics[width=\textwidth]{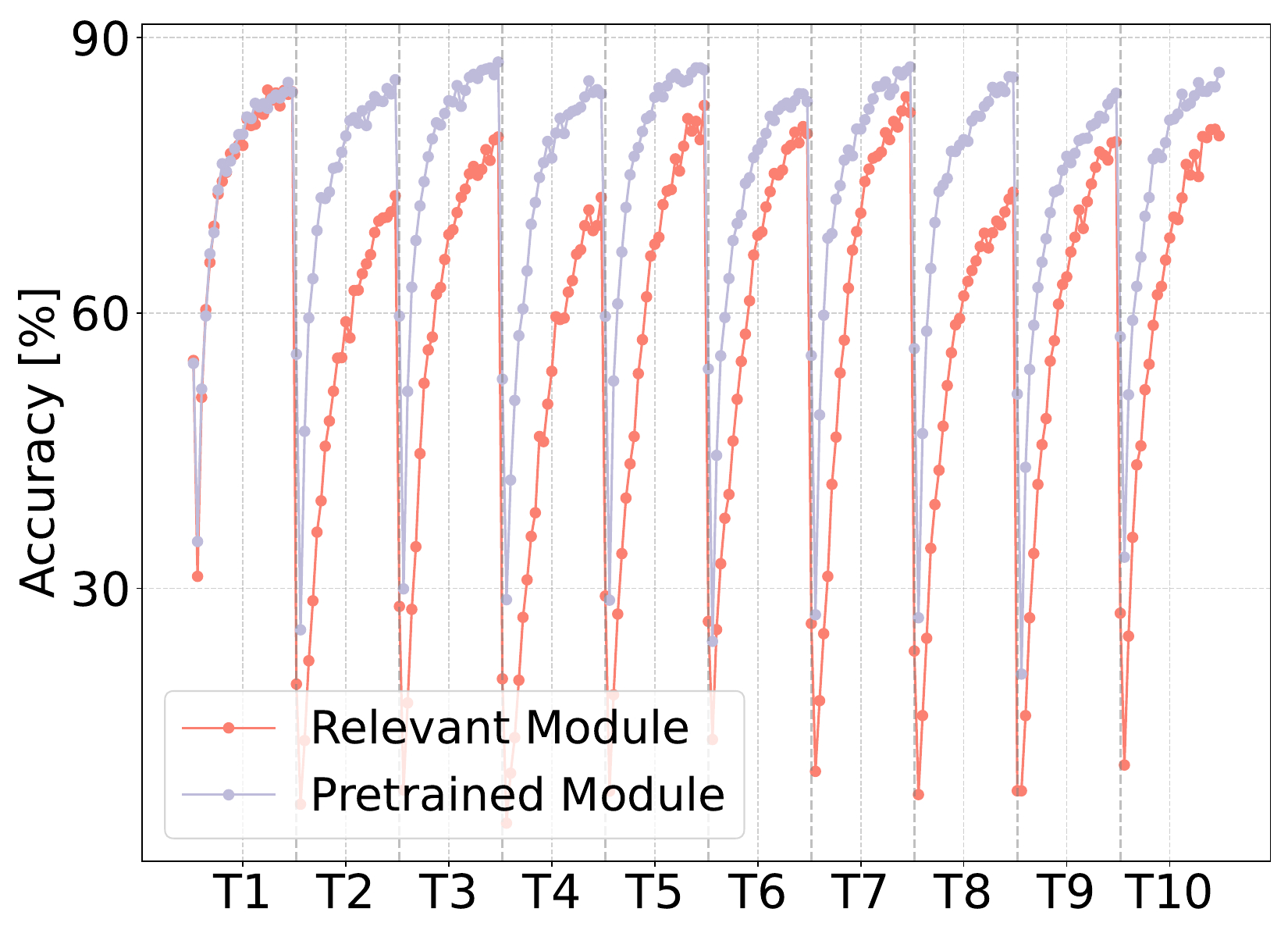}
        \caption{Per-task training accuracy}
        \label{knowledge_transfer_learning_curve}
    \end{subfigure}
    \hfill
    \begin{subfigure}{0.32\textwidth}
        \centering
        \includegraphics[width=\textwidth]{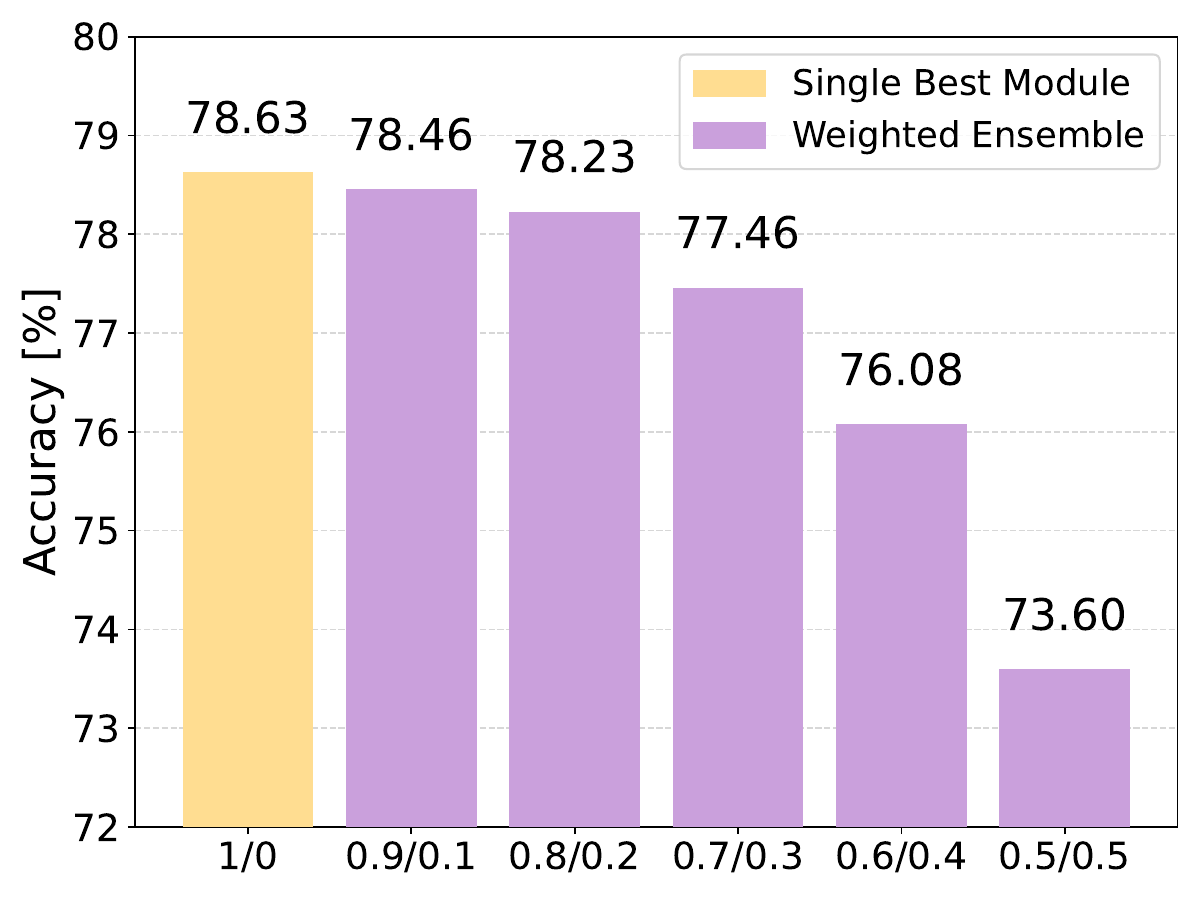}
        \caption{Ensemble performance}
        \label{ensemble_performance}
    \end{subfigure}
    \caption{Analysis of PAM across different evaluation settings. (a) Final incremental accuracy after all tasks, (b) per-task training accuracy illustrating the effect of initialization strategies for pruned adaptation modules, and (c) performance of weighted ensemble strategies compared to the single best module. Ratios show the relative contribution of the most confident module and the remaining modules (e.g., 0.9/0.1).}
    \label{fig:initialization_strategy}
\end{figure}

\paragraph{Pruning magnitude.} The pruning magnitude applied to the task-specific module is another critical factor. In this experiment, we vary the pruning magnitude applied to the task-specific module $\gamma_b$ across four settings: $0.95$, $0.96$, $0.97$, and $0.98$. Figure \ref{ablation_pruning} illustrates how the incremental performance changes as a function of the pruning level, providing insights into the trade-off between parameter reduction and predictive accuracy. Our analysis shows that introducing an extreme pruning level harms the performance and therefore we opt for an optimal pruning magnitude of $0.96$.

\paragraph{Module selection during inference.} Finally, we explore three strategies; two distance-based and one confidence-based for selecting the appropriate pruned adaptation module $\EuScript{S}_b$ at inference. In particular, the distance-based approaches, which compute the distance between the centroid of the test batch derived from the frozen backbone $\Phi$ and the stored centroids of previously encountered tasks, perform effectively on larger backbones ResNet101 and ResNet152. However, they fail to generalize well on smaller backbones ResNet18 and ResNet50. Consequently, we propose a refined, confidence-based approach that determines the correct pruned adaptation module $\EuScript{S}_b$ by leveraging the maximum softmax probability over the test batch. As shown in Figure \ref{ablation_test_time}, this confidence-based method consistently outperforms the distance-based strategies across all model sizes, ensuring more robust module selection at inference. Together, these findings offer empirical guidelines and highlight the efficiency of our design choices, and they provide a clear understanding of how each component contributes to the overall performance of our continual learning approach PAM.

\paragraph{Knowledge transfer between the modules.} Beyond our main ablations, we also conduct an additional investigation into the knowledge transfer dynamics within our PAM approach, as this aspect warrants further discussion. Specifically, we examine two different strategies for initializing newly added pruned adaptation modules: (i)  initializing from the most similar previously learned task (Relevant Module), and (ii) using pre-trained weights (Pretrained Module). As illustrated in Figure \ref{knowledge_transfer_barplot}, the `Pretrained Module' approach achieves a final accuracy of 77.4\% on ImageNet-R, outperforming the 72.4\% obtained by the `Relevant Module'. To understand this behavior we investigate the learning curves across all tasks in Figure \ref{knowledge_transfer_learning_curve} that shows the `Pretrained Module' strategy maintains a higher training accuracy. We hypothesize that pre-trained weights offer broader, more general feature representations, enabling more effective adaptation to new tasks. Reusing existing modules from other related tasks limits the model's ability to adapt to new task effectively, as their weights and feature spaces are already specialized for previous tasks. This makes it harder to align their representations with the current task compared to leveraging pre-trained general features, which offer a more flexible and transferable foundation. These findings emphasize the importance of selecting an appropriately general initialization strategy to promote stronger incremental learning performance.

\begin{figure}[t]
    \centering
        \captionsetup{font=small}
    \begin{subfigure}{0.32\textwidth}
        \centering
        \includegraphics[width=\textwidth]{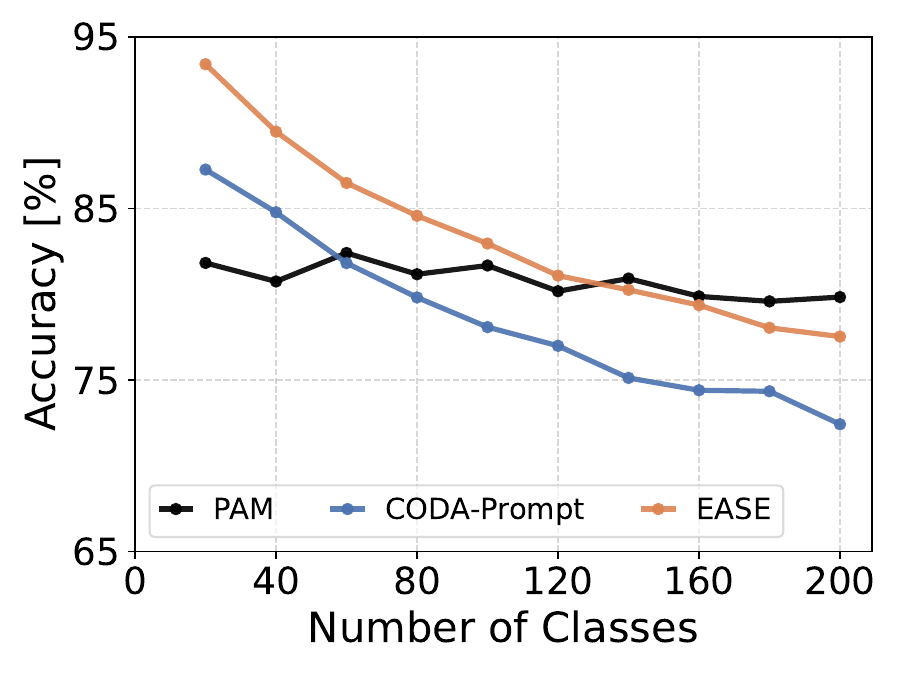}
        \subcaption{IN-R B0 Inc20}
    \end{subfigure}
    \begin{subfigure}{0.32\textwidth}
        \centering
        \includegraphics[width=\textwidth]{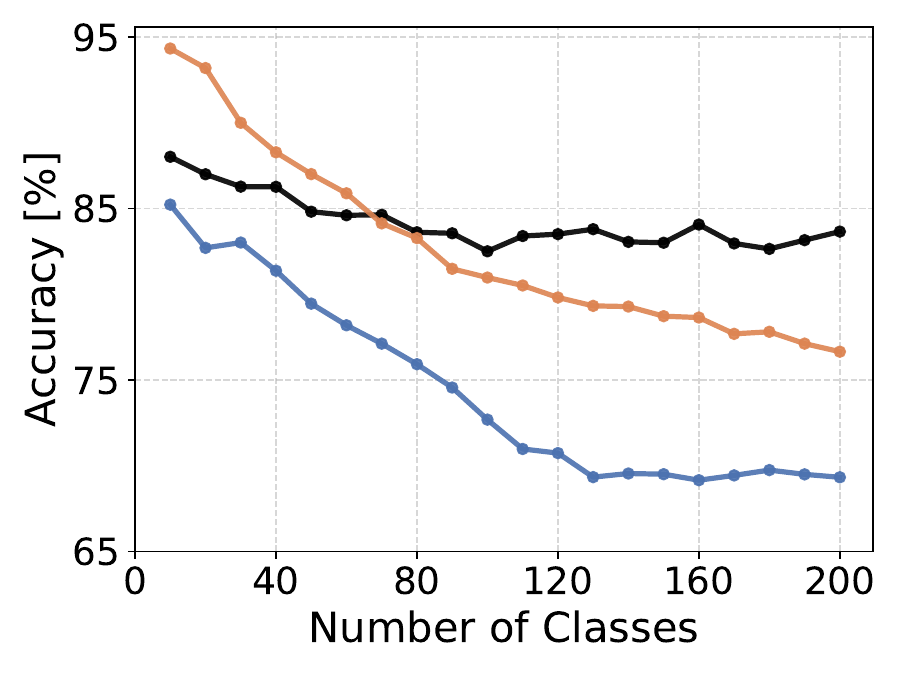}
        \subcaption{IN-R B0 Inc10}
    \end{subfigure}
    \begin{subfigure}{0.32\textwidth}
        \centering
        \includegraphics[width=\textwidth]{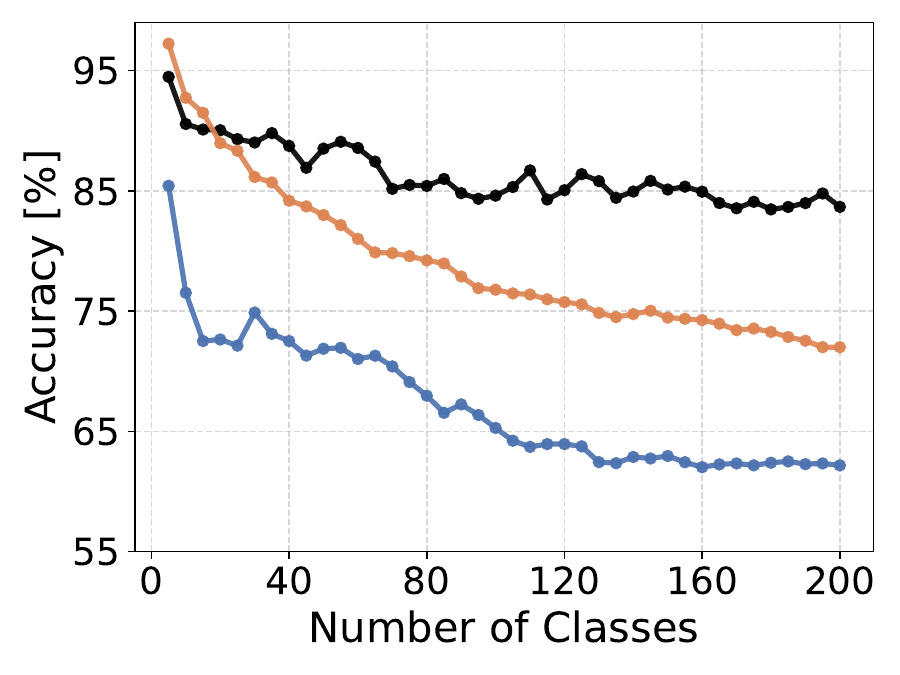}
        \subcaption{IN-R B0 Inc5}
    \end{subfigure}
    \caption{PAM achieves noticeably more stable accuracy throughout long-horizon experiments with challenging ImageNet-R (IN-R) dataset, reflecting its robustness against performance degradation as tasks accumulate.}
\end{figure}

\subsection{Discussion and Further Analysis}
\paragraph{Representation capacity.}
In our experiments, we evaluate various ResNet models and observe consistently strong final incremental accuracy. This is particularly noteworthy given that existing methods rely on much larger ViT backbone to reach high performance.
Despite this strength in final accuracy, PAM occasionally lags slightly in average accuracy compared to FM-based approaches. We attribute this difference to representational capacity where larger architectures tend to learn richer features. Specifically, PAM utilizes a ResNet backbone with fewer pre-trained parameters (3M-48M), whereas existing methods leverage larger pre-trained models (86M) that offer richer representational capacity. Indeed, when we scale up our model from pre-trained ResNet18 to pre-trained ResNet152, we observe improvements in both final and average accuracy. 
Overall, these findings illustrate that while large capacity backbones are more effective for adaptation but current FM-based approaches are falling short from that perspective since smaller ResNets can still offer competitive accuracy while being more efficient.

\paragraph{Longer learning sessions.}
Analyzing performance under longer learning sessions is critical in the continual learning paradigm, where models must remain robust as the number of tasks grows. To evaluate this, we conduct an extended study on the challenging ImageNet-R benchmark using varying numbers of tasks. We compare PAM against two strong state-of-the-art baselines representing different PEFT families: the prompt-based CODA-Prompt and the adapter-based EASE.
Across all experimental setups, PAM consistently exhibits more stable behavior as the number of classes increases. Although EASE begins with slightly higher initial accuracy, its performance deteriorates more rapidly over time, indicating weaker resilience in extended learning scenarios. CODA-Prompt suffers the steepest decline overall, highlighting the difficulty of maintaining prompt-based solutions under long sequences of tasks. In contrast, PAM maintains competitive accuracy throughout, demonstrating its suitability for realistic continual learning scenarios with slower performance decay and better long-term retention.

\begin{figure}[h]
\captionsetup{font=small}
  \centering
    \begin{subfigure}{0.37\textwidth}
    \includegraphics[width=\textwidth]{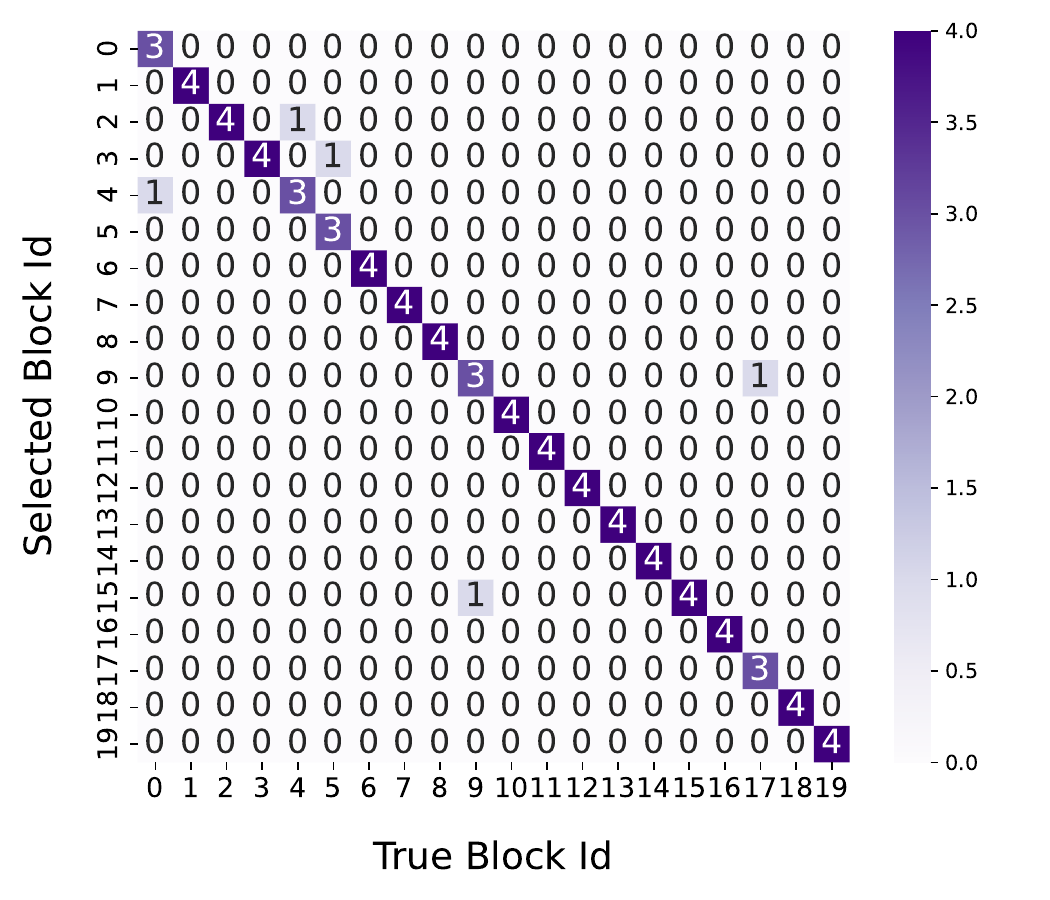}
    \caption{Module selection during inference}
    \label{fig:confusion_matrix}
  \end{subfigure}
    \hspace{1cm}
  \begin{subfigure}{0.37\textwidth}
    \includegraphics[width=\textwidth]{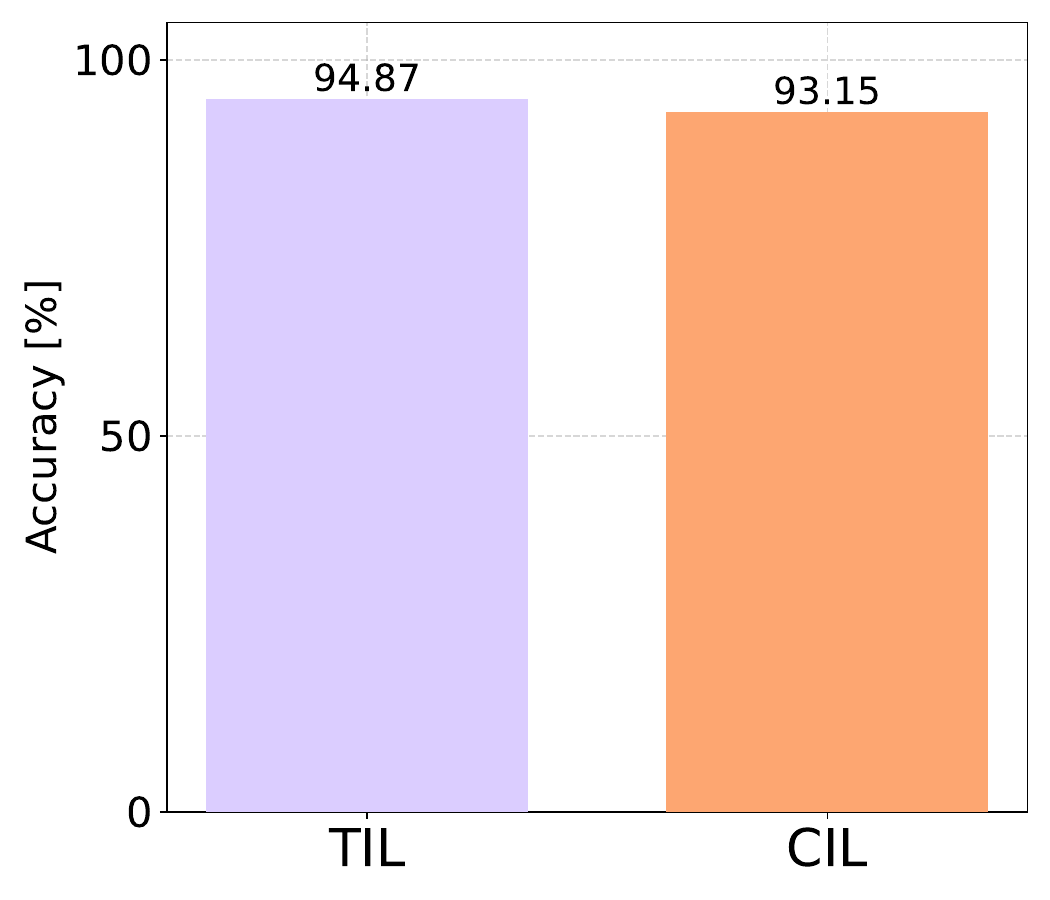}
    \caption{Incremental accuracy for TIL and CIL}
    \label{fig:til_cil}
  \end{subfigure}
\caption{Correct module alignment of the PAM:
(a) The confusion matrix shows that PAM reliably selects the correct adaptation module. (b) Comparing CIL and TIL results illustrates that providing true task identifiers yields a marginal improvement, indicating that PAM's performance is already close to the TIL upper bound.}
\vskip -0.1cm
\end{figure}

\paragraph{Proximity to the upper bound.}
To contextualize our results, we compare our performance in the challenging class-incremental learning setting against the task-incremental learning upper bound. Since task-incremental learning provides an idealized scenario with explicit task identities, this comparison highlights how close our approach gets to the best-possible performance.
Using the CIFAR B0 Inc5 setup, we measure how often the model chooses the appropriate module for each task. As depicted in Figure~\ref{fig:confusion_matrix}, the confusion matrix shows a strong diagonal pattern, indicating that the model consistently selects the correct module with only occasional mismatches. This reflects a high level of implicit task recognition and minimal confusion across tasks.
To understand how much this implicit selection affects overall performance, we also evaluate the system under a task-incremental learning setting, which serves as an upper bound because the correct module is explicitly provided for every test batch. As shown in Figure~\ref{fig:til_cil}, the gap between class-incremental learning and task-incremental learning accuracy is extremely small, demonstrating that PAM performs nearly as well as the ideal task-incremental learning scenario.

\paragraph{Ensemble strategy.}
Finally, we study different inference strategies, examining how to best leverage task-specific modules during prediction. In our default inference protocol, we rely on the prediction of the single most confident task-specific module, which consistently provides the strongest performance in our experiments. To reduce the reliance on a single module’s decision, we also explored an ensemble-based alternative. In this approach, we implemented a weighted ensemble strategy, assigning dominant weight to the most confident module while allowing the remaining modules to contribute proportionally through fixed ratios. This pilot study serves as an initial step toward assessing whether ensembles of task-specific experts can improve generalization. As reported in Figure \ref{ensemble_performance}, our findings show that naive ensembling does not outperform the single-module baseline, while using a principled weighting on ensembles shows greater promise.

\section{Conclusion}
\vspace{-5pt}
In this work, we introduce Pruned Adaptation Modules (PAM), a simple yet effective approach that bridges traditional CIL methods with emerging FM-based continual learning. Unlike recent strategies that depend heavily on large foundation models, PAM demonstrates that compact, pre-trained backbones such as ResNets can achieve competitive, or even superior, performance in CIL while dramatically reducing computational and storage costs. By freezing early layers and applying structured pruning to a lightweight, task-specific module, PAM achieves a $2$~-~$5\times$ reduction in trainable parameters and a $2$~-~$6\times$ smaller overall parameter footprint compared to state-of-the-art FM-based CIL methods.
Overall, PAM serves as a simple yet strong baseline for future FM-based continual learning research, highlighting that existing approaches may not be fully exploiting the powerful generalization capabilities of foundation models. Future work may explore extending PAM to transformer-based backbones, other continual learning scenarios, incorporating dynamic pruning strategies, or integrating PAM into larger FMs to combine efficiency with broad generalization capabilities.

\vspace{-5pt}
\section*{Broader Impact}
\vspace{-5pt}
Proposed approach reduces memory and computation requirements, enabling more accessible and energy-efficient continual learning research. As with any adaptive machine learning system, care should be taken to ensure responsible and fair deployment in real-world applications.

\vspace{-5pt}
\section*{Acknowledgements}
\vspace{-5pt}
This work is supported by the EU Horizon programme through SYNERGIES, a project under GA No. 101146542; and ELLIOT, a project under GA No. 101214398; and Dutch e-infrastructure with the support of SURF Cooperative using GA no. EINF-10242; and Turkish MoNE scholarship.

%\section*{Acknowledgements}
%This work is supported by; SYNERGIES, a project funded by the EU Horizon programme under GA No. 101146542; Dutch national e-infrastructure with the support of SURF Cooperative using grant no. EINF-4568, and Turkish MoNE scholarship.

\bibliography{reference}

\clearpage
\appendix
\section{Appendix}
\label{Appendix}
In this appendix, we present a supplementary strategy to our approach PAM for adaptively reusing pruned adaptation modules based on task similarity rather than defining entirely new ones for each novel task and demonstrate how this adaptive strategy influences overall model performance. We also provide a comprehensive explanation of the existing methods for both prompt- and adapter-based which are used in the evaluations and presented in the main paper, together with our pseudocode.

\subsection{Adaptive Approach for Initializing PAM}
In our original implementation, a new pruned adaptation module $\EuScript{S}_b$ was added for each incoming task. In this section, we try a more adaptive and efficient strategy that reuses existing pruned adaptation modules when a new task exhibits high similarity to previously encountered ones. This design yields a significantly more compact model by initializing fewer adaptation modules.

To quantify task similarity, we compute a task centroid $c_b$ by averaging the feature representations extracted from the shared frozen extractor $\Phi$ (i.e., the first three residual layers of a pre-trained ResNet) overall training samples in \(\mathcal{D}_b\), assuming that the mean feature representation effectively captures the overall data distribution. We measure the similarity between $c_b$ and the centroids of all previously encountered tasks $\{c_1, \dots, c_{b-1}\}$, using the Manhattan distance as in Eq \ref{eq:centroid}. 

\begin{equation}
d(\mathcal{D}_b, \mathcal{D}_i) \simeq \|c_b - c_i\|_1, \quad c_b = \frac{1}{n_b} \sum_{i=1}^{n_b} \Phi(x_i)
\label{eq:centroid}
\end{equation}

We then compute the average distance across all previously encountered tasks as in Eq. \ref{eq:avg_distance} and we scale it by a hyperparameter \(\beta\) which is a pre-determined hyperparameter to obtain a similarity threshold $\tau$ given in Eq. \ref{eq:threshold}.

\begin{equation}
\label{eq:avg_distance}
\bar{d} = \frac{1}{b} \sum_{b=1}^{b} d(\mathcal{D}_{b}, \mathcal{D}_i).
\end{equation}

\begin{equation}
\label{eq:threshold}
\tau = \beta \cdot \bar{d}.
\end{equation}

If the minimum distance \(d_{\min} = \min_{1\leq i < b} \|c_b - c_i\|_1\) between the \(c_b\) and any previous centroid \(c_i\) is less than \(\tau\), then the new task is deemed sufficiently similar to a previously encountered task. In such cases, rather than allocating a new pruned adaptation module, we reuse the module associated with the most similar task.
To mitigate the risk of catastrophic forgetting when reusing a module, we incorporate a knowledge distillation loss during the training, similar to LwF \cite{li2017learning}.

%\begin{equation}
%\label{eq:kdloss}
%\mathcal{L}_{\text{KD}} = \operatorname{KL}\Biggl( \log \operatorname{softmax}\Bigl(\frac{z_s}{T}\Bigr), \operatorname{softmax}\Bigl(\frac{z_t}{T}\Bigr) \Biggr).
%\end{equation}

%where \(z_s\) and \(z_t\) denote the logits generated by the student and teacher versions of the task-specific block, respectively, and \(T\) is the temperature parameter. Consequently, the overall loss is formulated as
%\begin{equation}
%\label{eq:kdloss+celoss}
%\mathcal{L} = \mathcal{L}_{\text{CE}} + \lambda \, \mathcal{L}_{\text{KD}}.
%\end{equation}
%where \(\mathcal{L}_{\text{CE}} = -\sum_{i=1}^{n_b} y_i^b \log p(y_i^b \mid x_i^b)\) is the cross-entropy loss computed over the new classes, and \(\lambda\) modulates the contribution of the distillation loss. In contrast, if no previous task is sufficiently similar (i.e., \(d_{\min} \ge \tau\)), a new task-specific block is initialized and standard cross-entropy loss is optimized as given in the main paper  Eq \ref{eq:celoss}.

The results presented in Figure \ref{fig:module_share} highlight a fundamental tradeoff between continual learning performance and parameter efficiency. By selectively reusing existing pruned adaptation modules, our approach reduces the total number of modules but this comes at the cost of diminished incremental accuracy. 

For instance, when the threshold hyperparameter \(\beta\) is set to $0.70$, Module $10$ is shared across Task $10$ and Task $19$ and yields 90.08\% accuracy. On the other hand, when \(\beta\) is set to $0.73$ it leads to more reusing across tasks where Module $3$ is used for Task $3$ and Task $11$ and Module $10$ for Task $10$ and Task $19$ with 76.16\% accuracy. Although it results lower in accuracy, we argue that developing such an adaptive mechanism is crucial for scalable continual learning. Our method  PAM provides a flexible framework for task reuse, making it straightforward to integrate into various applications.

\renewcommand\thefigure{A}
\begin{figure}[h]
\captionsetup{font=small}
    \centering
    \begin{subfigure}{0.4\textwidth}
        \centering
        \includegraphics[width=\linewidth]{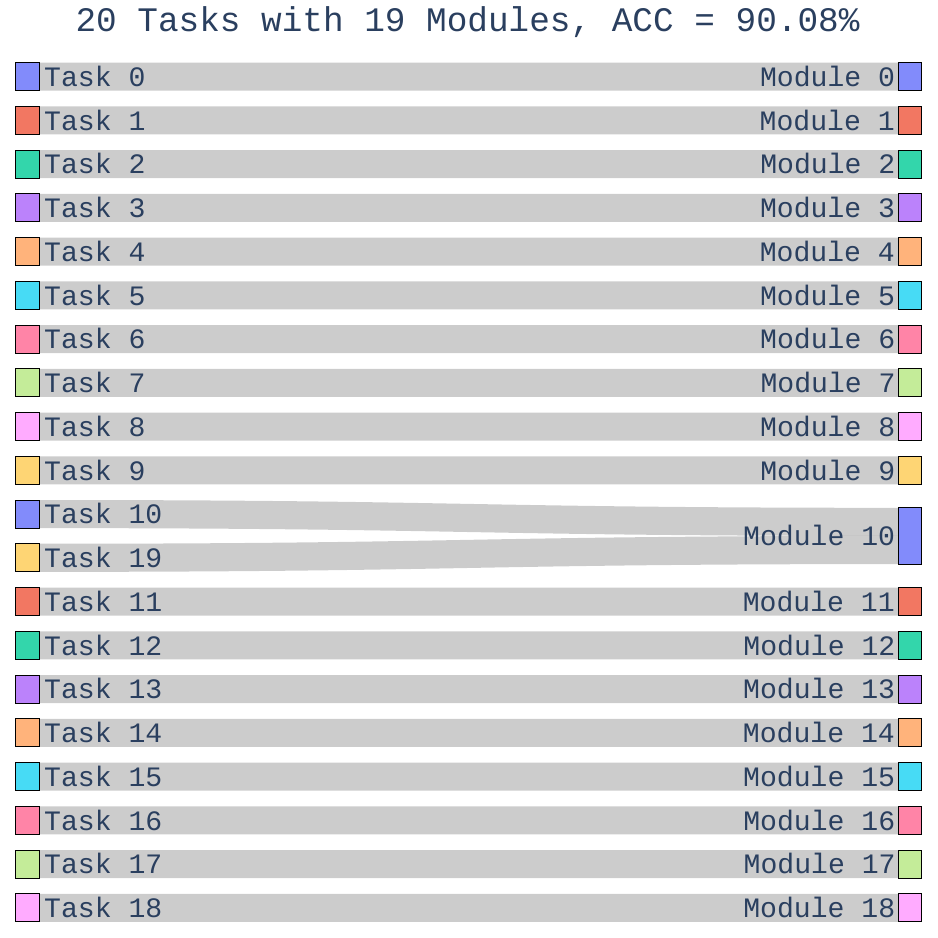}
        \caption{\(\beta\)=0.70}
        \label{fig:fig1}
    \end{subfigure}
    \begin{subfigure}{0.4\textwidth}
        \centering
        \includegraphics[width=\linewidth]{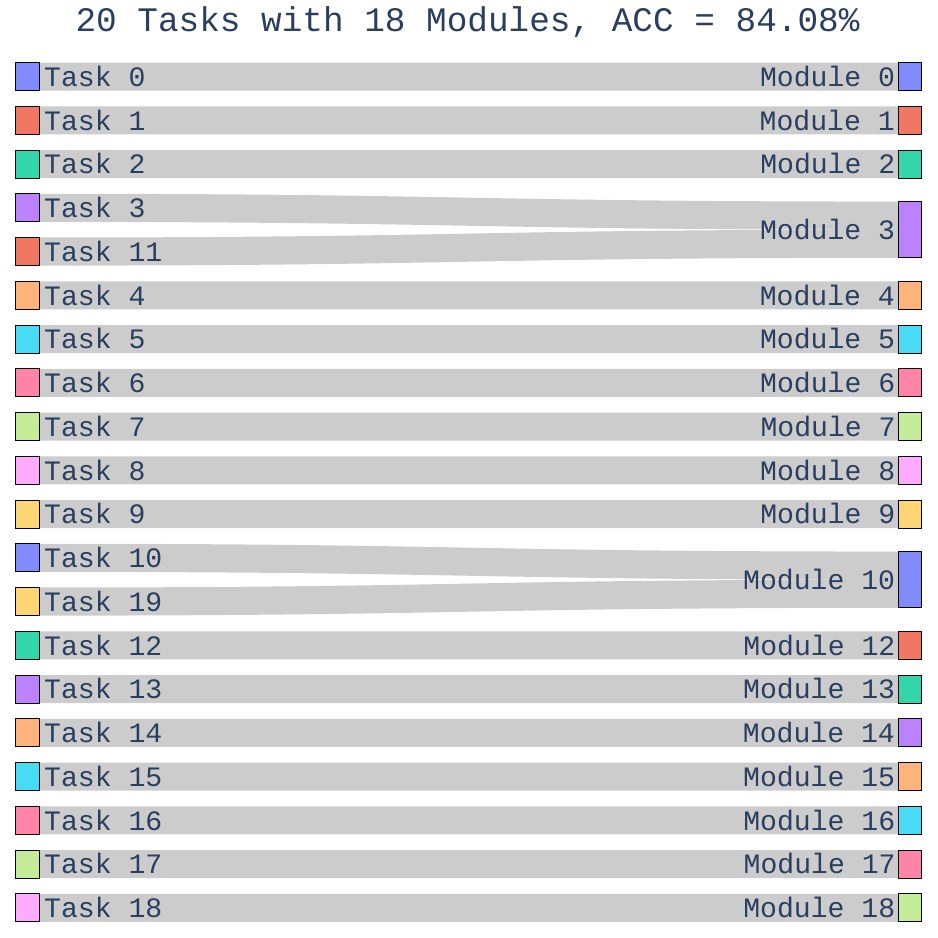}
        \caption{\(\beta\)=0.73}
        \label{fig:fig2}
    \end{subfigure}
    
    \vspace{0.5cm}  % Add space between rows

    \begin{subfigure}{0.4\textwidth}
        \centering
        \includegraphics[width=\linewidth]{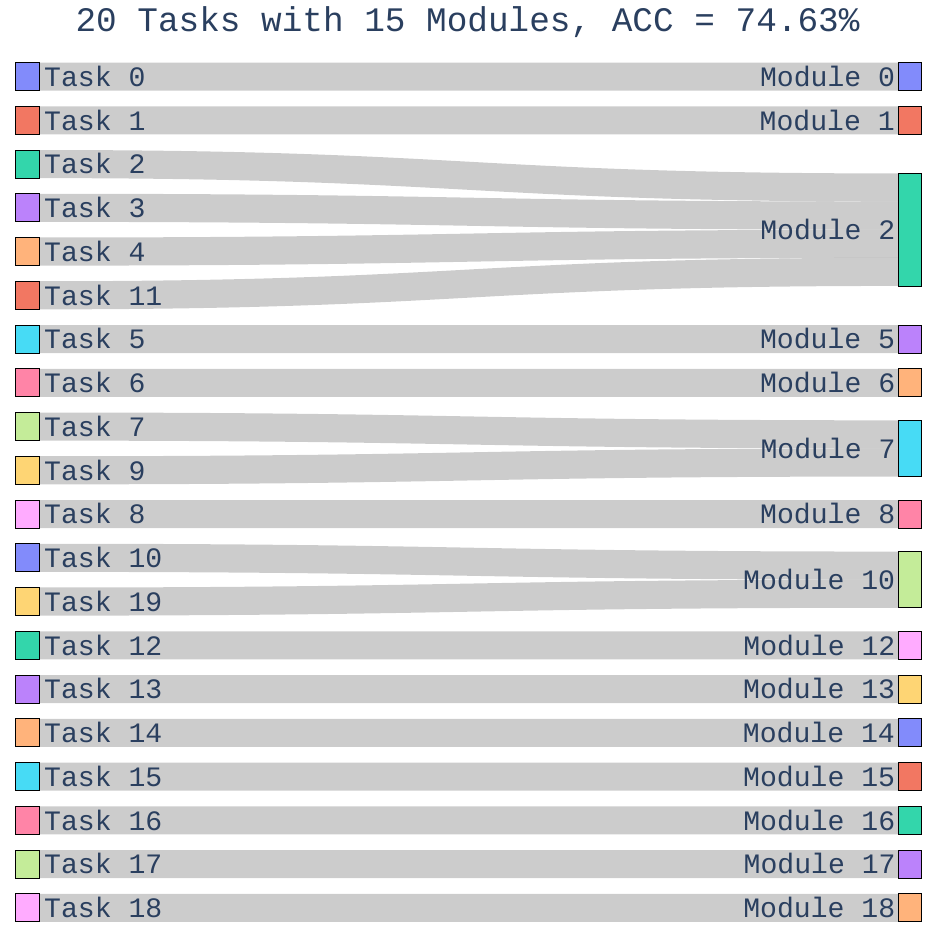}
        \caption{\(\beta\)=0.75}
        \label{fig:fig3}
    \end{subfigure}
    \begin{subfigure}{0.4\textwidth}
        \centering
        \includegraphics[width=\linewidth]{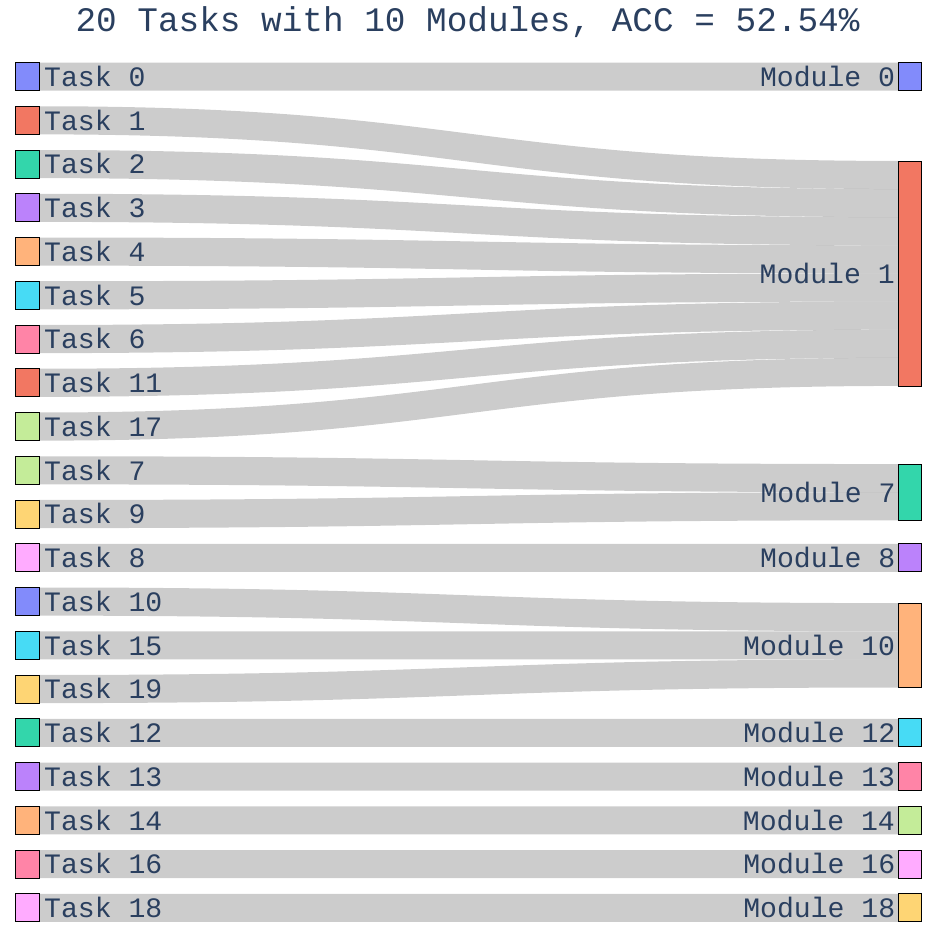}
        \caption{\(\beta\)=0.77}
        \label{fig:fig4}
    \end{subfigure}
    \caption{Analysis of the effect of threshold values on the adaptive initialization of pruned adaptation modules. As the threshold increases from (a) $0.70$ to (d) $0.77$, the model reuses existing modules more frequently, thereby reducing the total number of newly instantiated modules. }
    \label{fig:module_share}
\end{figure}

\newpage

\subsection{Details of Compared Methods and PAM}
\label{baseline-algo}
Below, we briefly describe the baseline methods we used to evaluate our approach. Then, we share the algorithm of PAM.

\begin{itemize}[leftmargin=*, align=left]
    \item \textbf{Finetune:} This baseline directly fine-tunes the pre-trained ViT backbone on each new task using the standard cross-entropy loss. While simple, it typically suffers from severe catastrophic forgetting as the entire model is updated without any constraints.
    
    \item \textbf{L2P \cite{l2p}:} L2P introduces pre-trained ViT into continual learning by freezing the backbone weights and using visual prompt tuning to capture features of new tasks. It constructs instance-specific prompts through a prompt pool organized via key-value mapping.
    
    \item \textbf{DualPrompt \cite{dualprompt}:} As an extension of L2P, DualPrompt refines the prompt mechanism by employing two types of prompts: general and expert prompts, while maintaining the instance-specific prompt construction process.
    
    \item \textbf{CODA-Prompt \cite{codaprompt}:} To overcome limitations in instance-specific prompt selection, CODA-Prompt replaces the selection process with an attention-based prompt recombination strategy, effectively eliminating the need for explicit prompt selection.
    
    \item \textbf{SimpleCIL \cite{simplecil}:} This method utilizes a vanilla pre-trained ViT model as initialization and builds a prototype-based classifier for each class, employing a cosine classifier for final prediction.
    
    \item \textbf{APER \cite{zhou2024revisiting}:} Extending SimpleCIL, APER aggregates the pre-trained and adapted models by treating the first incremental stage as the sole adaptation phase. This design unifies generalizability and task-specific adaptation within a single framework.

    \item \textbf{EASE \cite{ease}:} This method trains a distinct, lightweight adapter for each new task, thereby constructing task-specific subspaces. These subspaces enable joint decision-making while preserving prior knowledge, and a semantic-guided prototype complement strategy is employed to update old class features without requiring access to previous instances.
    
\end{itemize}

\begin{algorithm}
\caption{PAM: Pruned Adaptation Modules}
\label{alg:cil}
\begin{algorithmic}[1]

\State \textbf{\underline{Training Phase:}}
\For{each task $D_b \in \{D_1, D_2, \ldots, D_B\}$}
\For{$e$ in range epoch}

\If{$e = 1$}
%\State Compute task centroid $c_b$ using Eq.~\ref{eq:centroid} 
\State Train $\gamma_b$ using the loss function in Eq.~\ref{eq:celoss}
\State Rank saliency of the filters in $\gamma_b$ using Eq.~\ref{eq:l1norm}
\State Obtain $\EuScript{S}_b$ with the saliency-based pruning
\Else
\State Train $\EuScript{S}_b$ with remaining filters using Eq.~\ref{eq:celoss}
\EndIf

\EndFor
\EndFor

\medskip
\State \textbf{\underline{Inference Phase:}}
\For{each test batch $\mathbf{x}_{test}$}
    \State Get class probabilities for each $\EuScript{S}_b$ using Eq. \ref{eq:prob_dist}
    \State Select the most confident $\EuScript{S}_{\hat{b}}$ using Eq. \ref{eq:confidence}
    \State Get predictions for $\mathbf{x}_{test}$ using Eq. \ref{eq:logits_sam} with $\EuScript{S}_{\hat{b}}$
\EndFor

\end{algorithmic}
\end{algorithm}

\renewcommand\thefigure{B}
\begin{figure*}[h]
\captionsetup{font=small}
\centering
\begin{subfigure}{0.33\textwidth} % slightly less than 1/4
    \includegraphics[width=\linewidth]{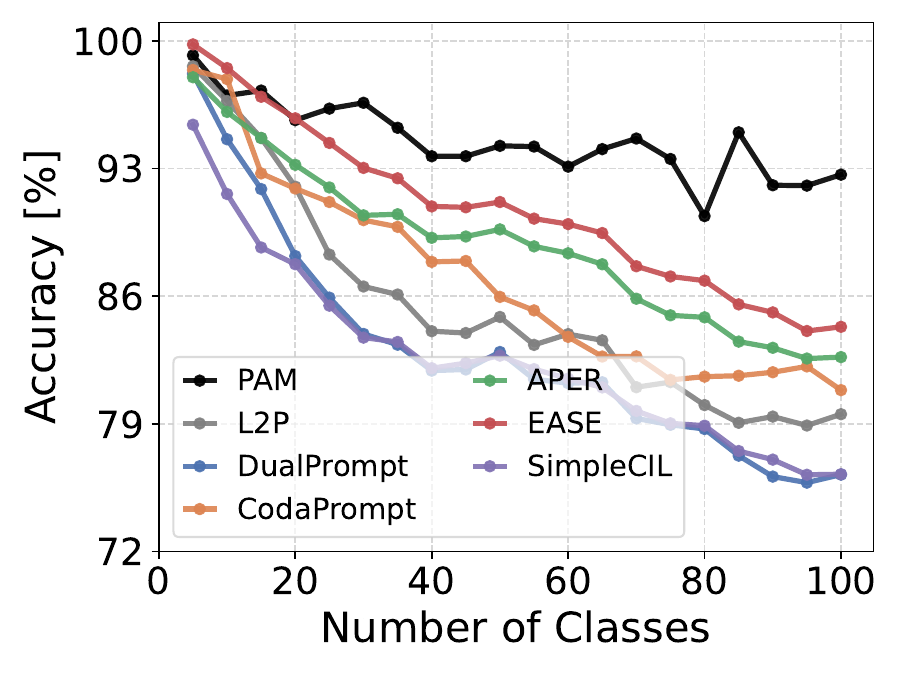}
    \caption{CIFAR B0 Inc5}
\end{subfigure}%
\begin{subfigure}{0.33\textwidth}%
    \includegraphics[width=\linewidth]{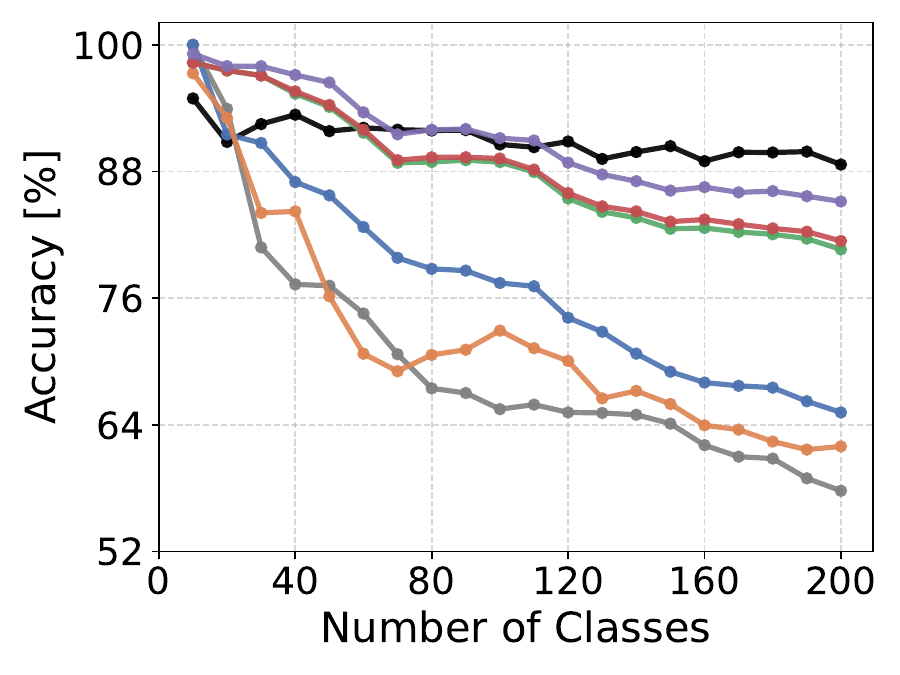}
    \caption{CUB B0 Inc10}
\end{subfigure}%
\begin{subfigure}{0.33\textwidth}%
    \includegraphics[width=\linewidth]{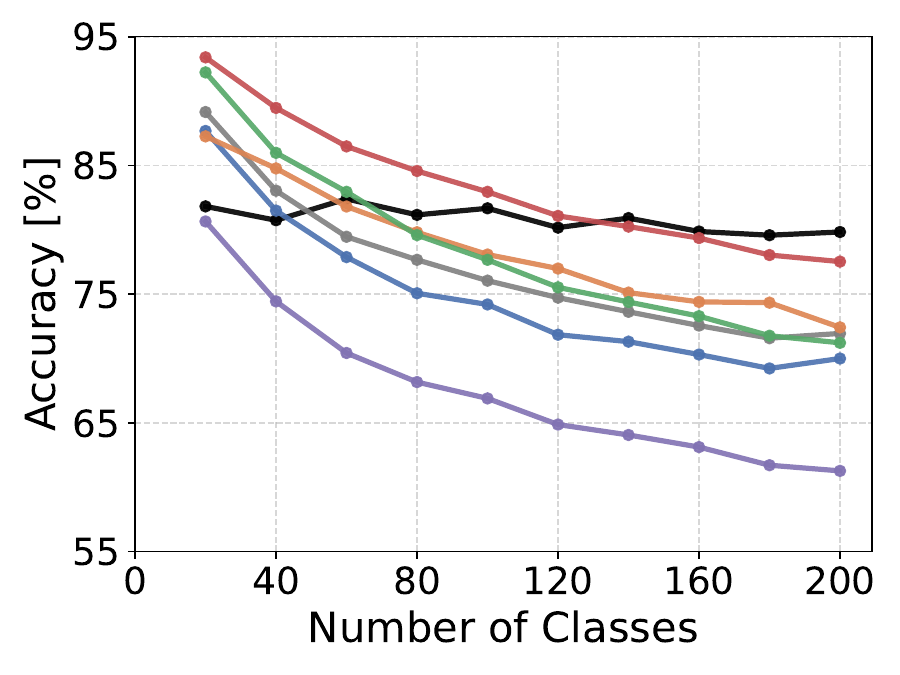}
    \caption{ImagenetR B0 Inc20}
\end{subfigure}%
%\begin{subfigure}{0.25\textwidth}%
%    \includegraphics[width=\linewidth]{figs/lineplot_cars.pdf}
%    \caption{CARS B0 Inc10}
%\end{subfigure}
\caption{Incremental accuracy flow of each method over sequential tasks where comparison methods utilize ViT-B/16 and PAM employs ResNet152. PAM maintains a more stable SOTA performance across tasks.}
\label{fig:line_plots}
\end{figure*}

%%%%%%%%%%%%%%%%%%%%%%%%%%%%%%%%%%%%%%%%%%%%%%%%%%%%%%%%%%%%

\end{document}